\documentclass[letterpaper, 10 pt, conference]{ieeeconf}
\IEEEoverridecommandlockouts
\usepackage[T1]{fontenc}
\usepackage{balance}

\IEEEoverridecommandlockouts

\usepackage{cite}
\usepackage{graphicx}
\usepackage{url}
\usepackage[most]{tcolorbox}
\usepackage[latin1]{inputenc}
\usepackage{soul}
\usepackage{multirow}
\let\labelindent\relax
\usepackage{enumitem}
\usepackage{pbox}
\usepackage{amsmath,amssymb,amsfonts,siunitx}
\usepackage{amsthm}
\usepackage{algorithmic}
\usepackage{algorithm}
\usepackage{bm}
\usepackage{graphbox}
\usepackage{textcomp}
\usepackage{subcaption}
\usepackage[dvipsnames]{xcolor}
\usepackage[export]{adjustbox}
\usepackage{booktabs}
\usepackage[labelfont=bf,font=footnotesize]{caption}

\usepackage{custom_fmt}
\usepackage{microtype}
\makeatletter
\let\NAT@parse\undefined
\makeatother

\usepackage{hyperref}
\hypersetup{
colorlinks=true,
linkcolor=black,
citecolor=black,
urlcolor=blue,
}
\usepackage{cleveref}
\crefname{problem}{Problem}{Problems}
\crefname{figure}{Fig.}{Figures}

\title{\LARGE\bf Prompting Robot Teams with Natural Language}

\author{
Eduardo Sebasti\'{a}n$^{1,*}$ 
\and Nicolas Pfitzer$^{1,2,*}$
\and Ajay Shankar$^{1}$
\and Amanda Prorok$^{1}$%
\thanks{$^1$Department of Computer Science and Technology, University of Cambridge, UK; e-mails: \texttt{\small \{es2121, np632, as3233, asp45\}@cst.cam.ac.uk}. $^2$Department of Mechanical and Process Engineering, ETH Zurich, Switzerland. *Equal contribution.}%
\thanks{This work was supported in part by ARL DCIST CRA W911NF-17-2-0181 and by a Leverhulme Trust Research Project Grant. We gratefully acknowledge their support.}%
}

\newcommand\copyrighttext{%
  \footnotesize \textcopyright This paper has been accepted for publication at IEEE Robotics and Automation Letters. Please refer to the official manuscript for citing.}
\newcommand\copyrightnotice{%
\begin{tikzpicture}[remember picture,overlay]
\node[anchor=south,yshift=10pt] at (current page.south) {\fbox{\parbox{\dimexpr\textwidth-\fboxsep-\fboxrule\relax}{\copyrighttext}}};
\end{tikzpicture}%
}

\begin{document}
\maketitle
\copyrightnotice

%%%%%%%%%%%%          
% ABSTRACT %
%%%%%%%%%%%%

\begin{abstract}
This paper presents a framework to prompt multi-robot teams with high-level tasks using natural language expressions. Our objective is to use the reasoning capabilities of language models in understanding and decomposing multi-robot collaboration and decision-making tasks, but in settings where such models cannot be called at deployment time. However, it is hard to specify the behavior of an individual robot from a team instruction, and have it continuously adapt to actions from other robots. This necessitates a framework with the representational capacity required by the logic and semantics of a task, and yet supports decentralized, real-time operation. We solve this dilemma by recognizing that a task can be represented as a deterministic finite automaton, and that recurrent neural networks (RNNs) can encode numerous automata. This allows us to \textit{distill} the logic and sequential decompositions of sub-tasks obtained from a language model into an RNN, and align its internal states with the semantics of a given task. This leads to a tiny model that encapsulates the reasoning of the language model and can be implemented onboard.
To interpret the internal state of the RNN for a decentralized execution, we train a graph neural network control policy conditioned on the hidden states of the RNN and the language embeddings.
We present evaluations on simulated and real-world multi-robot tasks that require sequential and collaborative behavior by the team, demonstrating scalable, robust, real-time performance ---
\texttt{sites.google.com/view/prompting-teams}.
\end{abstract}

%%%%%%%%%%%%%%%%           
% INTRODUCTION %
%%%%%%%%%%%%%%%%

\section{Introduction}\label{sec:intro}
\begin{figure}[!t]
    \centering
    \includegraphics[width=0.95\linewidth]{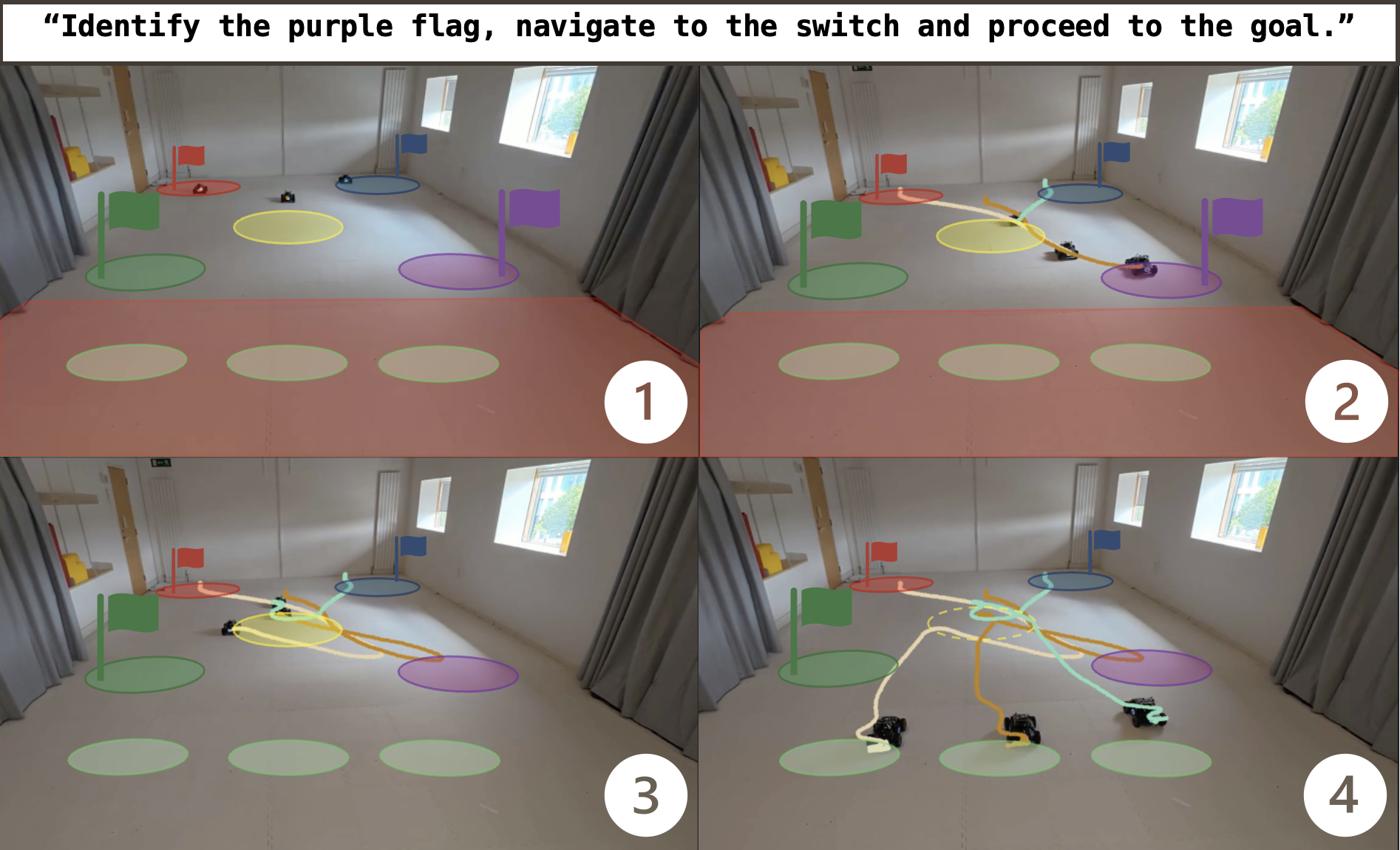}
    \caption{A deployment of our method on RoboMasters \cite{blumenkamp2024cambridge}. The robots are tasked with identifying and retrieving the purple flag. Once the flag has been placed on the yellow switch, the access to the red part of the arena is activated, and the robots navigate to their respective final destinations.}
    \label{fig:hero}
\end{figure}

Teams of robots have a natural application in scenarios that benefit from strength in numbers \cite{hartmann2022long, edwards2023collaborative, kim2023multi, liu2024multi}.
When tasks involve robots supporting human operators, we anticipate the operators to provide abstract commands for the task (e.g., ``explore this zone and find the missing people''), and supply semantically relevant hints (e.g., ``there are likely ten people on the other side of the damaged bridge'').
Given such information, human teams excel at collaborating and reasoning through language by
decomposing, distributing and sequencing sub-tasks, which is a highly sought-after capability for teams of robots.
The goal of this paper is to develop a single, light-weight model that enables decentralized coordination and reasoning in teams of robots commanded through natural language specifications (Fig. \ref{fig:hero}).

A fundamental challenge in realizing such multi-robot autonomy is the gap between the \textit{natural-language} {expression of a task} and its representation as an \textit{algorithm} that robots can execute \cite{cohen2024survey}.
Early work involved extensive manual design
of%
one-to-one correspondences with algorithmic primitives~\cite{bugmann2004corpus, kollar2010toward}, and thus had limited reasoning capabilities.
In contrast, language models today can side-step such design by encapsulating vast amounts of human language data.
Large multi-modal and language models (LMMs/LLMs) have recently shown excellent reasoning, disambiguation and even planning for single- and multi-robot tasks~\cite{biggie_tell_2023, li_language_2024}.
However, their computational
demands still make them inaccessible to practical, deployable robot platforms.
While some recent work has developed methods to effectively  extract 
capabilities from these models into more succinct representations \cite{ravichandran_distilling_2025, morad_language-conditioned_2024, qu_learning_2025, zhan_latmos:_2025}, they depend on an infrastructure that can be hard to adapt to real-time, dynamic and unforeseen settings.

In this work, we bridge the gap between high-level reasoning and its real-time decentralized multi-robot implementation.
We leverage the observation that robot tasks are \textit{algorithms}: algorithms can be represented using automata, which can, in turn, be encoded by recurrent neural networks (RNNs).
We train this RNN offline using task decompositions obtained from an LLM, distilling its reasoning capabilities and aligning its internal states with the tasks' semantics. 
This leads to a single, tiny RNN that encapsulates a variety of multi-robot tasks in its internal state.
This internal state is then used to condition a graph neural network-based policy trained using multi-agent reinforcement learning, thus connecting the reasoning process 
to a lower-level decentralized execution of the sub-tasks in a feedback loop.
We evaluate this via simulations and real experiments on multi-robot problems that require reasoning and sequential decision-making.
Our results demonstrate a first in achieving onboard, real-time, decentralized multi-robot reasoning and collaboration---all from a natural language command.

\section{Background}\label{sec:background}
Early work on interpretable natural-language commands \cite{bugmann2004corpus, levit2007interpretation, macmahon2006walk, kollar2010toward, howard_intelligence_2022} focused on developing ontologies with one-to-one correspondences with algorithmic primitives.
However, this requires intense manual design and cannot accommodate ambiguous intent. % and reasoning.
LLMs \cite{brown2020language, wu2023tidybot, ravichandran_spine:_2025} and LMMs \cite{o2024open,team2025gemini, yu_co-navgpt:_2025} overcome this by encapsulating
{diverse} human knowledge and style into one model.
Hence, they exhibit effective `reasoning' and flexibility to handle ambiguous yet expressive language prompts in both single- and multi-robot problems \cite{biggie_tell_2023, li_language_2024, rajvanshi_saynav:_2024, han_large_2024}.
For instance, LLMs can be used to enable algorithmic tasks grounded in semantics \cite{cladera_air-ground_2025}, or create formations \cite{venkatesh_zerocap:_2025} hard to specify in a metric space.
Although effective, these works assume an infrastructure to connect the robots with the LLM/LMM in real-time.

To alleviate this, recent work has developed computationally efficient alternatives that rely on \textit{distillation}, and distributed coordination~\cite{ravichandran_distilling_2025, morad_language-conditioned_2024, qu_learning_2025, zhan_latmos:_2025, nematollahi_lumos:_2025}.
For instance,%
{it is possible to distill an LLM} into a smaller model that captures the relevant skills for robot task decomposition~\cite{ravichandran_distilling_2025}; however, the decomposition cannot integrate feedback in real-time.
Meanwhile, solutions using reinforcement learning allow onboard deployment but do not handle team-level prompts or credit assignment \cite{morad_language-conditioned_2024}.
{Yet another alternative is language translation:}
LLMs have shown surprising skills in translating human commands into linear \cite{dai_optimal_2024} or signal temporal logic \cite{chen2024autotamp}, or planning domain definition language \cite{strader_language-grounded_2025} formulas.
However, formal-based approaches need an initial time-consuming offline phase for each formula to be solved.
Besides, they do not generate the low-level policies that actually \textit{solve} the sub-tasks or feed back to the reasoning module to handle unexpected phenomena.
Achieving decentralized `in-the-loop' execution starting from natural, team-level commands is, thus, an open challenge.

\textbf{Our solution} overcomes these by distilling LLM-generated sub-task logic into a single tiny RNN, drawing on the connections between automata theory and recurrent models \cite{li2024connecting}.
By employing a graph-based low-level RL policy for each sub-task, we enable both decentralization, and real-time feedback that effects transitions in the automata.

%%%%%%%%%%%%%%%%%%%           
%% PRELIMINARIES %%
%%%%%%%%%%%%%%%%%%%

\section{Preliminaries}\label{sec:prosta}

In this work, a team of $\mathsf{N}$ robots is given a task as a natural-language instruction $\mathcal{W}$ (e.g., ``Guide all people back to base''), which is typically a composition of some logically distinct `sub-tasks', $w$ (e.g., ``search wide'', ``navigate group to base'', \ldots).
The team has local connectivity, formalized as a time-varying, undirected graph $\mathcal{G}_k = ({V}, {E}_k)$, where ${V} = \{1, \hdots, \mathsf{N}\}$ is the set of robots and ${E}_k \subseteq {V} \times {V}$ is the set of edges.
We define the set of neighbors of robot $r$ at time $k$ as ${N}_{r,k}$. 
The objective is to train control policies that complete $\mathcal{W}$, relying only on information available through $\mathcal{G}_k$.
Solving $\mathcal{W}$ involves reasoning and solving sub-tasks sequentially, which we formalize as a Deterministic Finite Automata ({DFAs}).
To learn to solve each sub-task effectively, we assume that only a reward signal is available, which we formalize as a Decentralized POMDP (Dec-POMDP).

\textbf{Deterministic Finite Automata.}
A DFA is a tuple \mbox{\(\mathcal{A} = (\mathcal{H}, \Sigma, \delta, h_0, \mathcal{F})\)}, where
\(\mathcal{H}\) is a finite set of sub-tasks to be solved;
\(\Sigma\) is a finite alphabet of inputs, with each proposition \(p \in \Sigma\) a boolean that describes an event that can trigger a transition between states;
\(\delta: \mathcal{H} \times \Sigma \to \mathcal{H}\) is a transition function such that \mbox{\(h_{k+1} = \delta(h_k, p_k)\)}, with \(h_k, h_{k+1} \in \mathcal{H}\) and input \(p_k \in \Sigma\);
\(h_0\in \mathcal{H}\) is an initial state; and,
\(\mathcal{F} \subseteq \mathcal{H}\) is the set of accepting states modelling the completion of the task.
Note that individuals in a multi-robot team completing the same task may step through $\mathcal{A}$ separately.
We develop $\mathcal{A}$ as a computational representation of $\mathcal{W}$, and thus its states, $h \in \mathcal{H}$, are representations of sub-tasks $w$.
The typical challenge is that $\mathcal{A}$ is unknown, and constructing it from $\mathcal{W}$ involves deduction and logic reasoning.
While the sub-tasks may be describable as Markov processes, switching between them requires memory, captured by DFAs.

\textbf{Decentralized POMDP.}
Given a DFA, the team must also be able to collaborate and complete the sub-tasks within it.
We define each sub-task $h \in \mathcal{H}$ as a Dec-POMDP
\mbox{$\mathcal{M}_h \kern -0.1cm = ({V}, {S}, \{A_r\}_{r\in V}, T_h, R_h, \{\Omega_r\}_{r \in V}, O, b_h, \gamma_h)$},
with
a finite set of robot states $S$, and actions $\{A_r\}_{r\in V}$, such that \mbox{$A = \times_{r \in V} A_r$} is the joint action space.
Functions \mbox{$T_h = P(s_{k+1}=s^{\prime}|h, s_k=s, a_k=a)$} and \mbox{$R_h: S\times A \rightarrow \mathbb{R}$} describe the dynamics and the reward of the sub-task $h$.
Observations are described by the finite set $\{\Omega_r\}_{r\in V}$.
Finally, the function $O = P(o_{k+1}=o|s_{k+1}=s^{\prime}, a_k=a)$ describes the measurement model of the team, $b_h$ defines an initial probability distribution of the sub-task, and $\gamma_h$ is a discount factor. 
In a Dec-POMDP, the goal is to find a policy $\pi_h$,
{
\setlength{\abovedisplayskip}{1.1ex}
\setlength{\belowdisplayskip}{1.1ex}
\[
\pi_h = \arg \max_{\pi} \;\;
\mathbb{E}\left[\sum_{k=0}^{\infty} \gamma_h R_h | b_h, \pi\right],
\]
}
such that the sub-task $h$ is solved.

\textbf{Problem formulation.}
Our objective is to bridge the gap between high-level reasoning and low-level multi-robot execution by
\textit{(i)} obtaining a mapping from natural language instructions, $W$, into a structured computational representation (i.e., a DFA),
and then,
\textit{(ii)}, learning a control policy that interprets the states of this representation (i.e., the sub-task sequence) to solve the overall mission.
This solution must be decentralized, allowing each robot to utilize real-time feedback from its local environment and neighbors to progress through the automaton asynchronously.

\begin{problem}\label{prob:prob}
Find a policy $\pi$ that solves the task specified by natural-language specification $\mathcal{W}$ and modeled by the set-product \mbox{$\mathcal{A} \times \{\mathcal{M}_h\}_{h \in \mathcal{H}}$}, such that $\pi$ relies only on the information available through $\mathcal{G}_k \hbox{ } \forall k$. 
\end{problem}

We recognize (and indeed use) the fact that LLMs/LMMs today broadly possess the representational capacity to solve \Cref{prob:prob} by, for instance, autoregressively generating action sequences for each robot at each time-step, adapting them with new information.
This requires persistent low-latency connection to servers, which is wasteful, uneconomical and often infeasible.
Our objective instead is to produce a small neural network $\pi$ that implements a solution to $\mathcal{W}$ as a robust real-time feedback system easily distributed onto each robot.

%%%%%%%%%%%%%%           
%% SOLUTION %%
%%%%%%%%%%%%%%
\section{Automata-based Reasoning and Collaboration for Robot Teams}\label{sec:technical_approach}
\begin{figure}
    \centering
    \includegraphics[width=\linewidth]{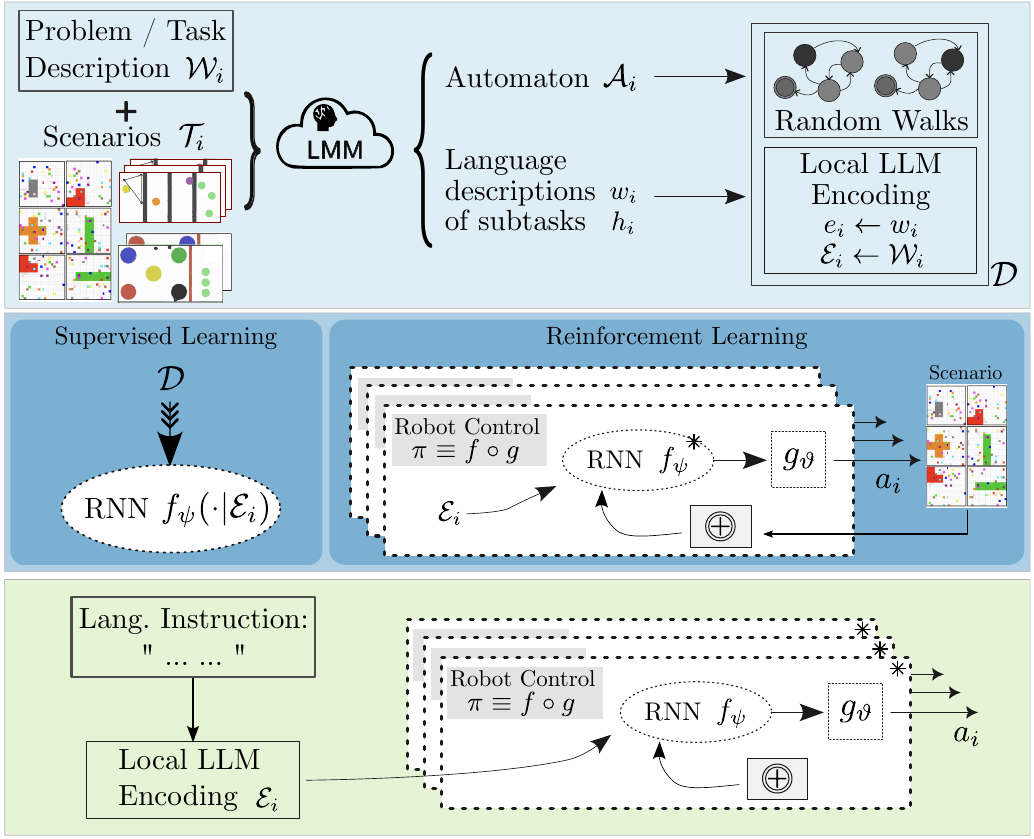}
    \caption{{\setlength{\fboxsep}{.5pt}
    Overview of our framework.
    An \fcolorbox{phase1}{phase1!35}{offline generation} phase compiles a dataset $(\mathcal{D})$ composed of sequences of sub-tasks and corresponding language embeddings for \emph{multiple tasks}, {obtained by prompting an LMM with diverse task specifications.}
    Next, two \fcolorbox{phase2}{phase2!25}{offline training} stages follow: (left) an RNN is trained with a supervised loss, using $\mathcal{D}$, to learn a set of semantic-aware automata; (right) a GNN-based policy is trained with an RL loss, using the frozen ($^\ast$) RNN to learn a low-level policy that solves all sub-tasks by interpreting their latent representation in the RNN.
    Finally, during \fcolorbox{phase3}{phase3!25}{online deployment}, robots only have to translate the language command into an embedding, and execute the RNN and GNN to complete the task.
    }
    }
    \label{fig:overview}
\end{figure}

To solve Problem \ref{prob:prob}, we propose a novel framework that works in three steps. 
First, an automated dataset generation phase queries an LMM with multi-robot tasks/problems, and compiles a dataset of the logic and sub-tasks needed to solve them.
Second, a supervised learning phase distills the reasoning of the LMM by training an RNN that encapsulates the desired collection of automata in the dataset.
Third, a reinforcement learning phase learns decentralized control policies that interpret the internal state of the RNN and solve the sub-tasks encoded in that internal state.
Fig. \ref{fig:overview} depicts a schematic of the proposed methodology, which we now elaborate in the following three subsections.

\subsection{Automatic generation of reasoning data}\label{subsec:generation}

Our first step is to generate a dataset $\mathcal{D}$ that captures diverse tasks and their underlying logic.
We will denote the set of tasks as $\mathcal{T}$, with $i \in \mathcal{T}$ a task.
To generate $\mathcal{D}$, we need a method that is conducive to learning the relationship between the language specification of a task and its representation as a set of sub-tasks and events in the automata.

For each task $i$, we first prompt an LMM with a task description.
The LMM then generates a formal decomposition for each of these tasks as scripts. For each task in $\mathcal{T}$, we use its formal decomposition to perform rollouts as random walks and collect $\mathsf{L}$ sequences of length $\mathsf{K}$ comprised of sub-tasks and the events that triggered the transitions between sub-tasks.
We represent sub-tasks as one-hot vectors $h^l_{i,k}$, their corresponding sentence language embedding $e^l_{i, k}$, and the atomic propositions $p^l_{i, k}$ that represent the events as one-hot vectors.
We also collect $\mathcal{E}^l_i$, the sentence language embedding associated to the high-level natural-language command $\mathcal{W}^l_i$.
The dataset is a collection of successful steps that fulfill the tasks, $\mathcal{D}= \{\{\mathcal{E}^l_i, \{h^l_{i, k}, e^l_{i, k}, p^l_{i, k}\}_{k=1}^{\mathsf{K}}\}_{l=1}^{\mathsf{L}}\}_{i\in\mathcal{T}}.$

As shown in \Cref{fig:overview} (top), we automate the dataset generation by grounding the natural-language specifications $\mathcal{W}^l_{i}$ in the actual environment that will be used to train the reinforcement learning policies.
Instead of manually handcrafting each $\mathcal{W}^l_{i}$, we procedurally generate $\mathsf{L\cdot{}M}$ scenarios of the tasks to be solved.
We then feed their features, including visual information, to the LMM with a prompt that requests natural-language specifications $\mathcal{W}^l_{i}$ that match the scenarios.

Detailed examples of prompts and outputs of the LMM models can be found on the \href{https://sites.google.com/view/prompting-teams}{website}.
This process not only automates the data generation step, but also grounds it in the actual environment that the robots will operate in. The dataset generation process is conducted offline and only once. We now detail the process by which a neural network can be trained to \textit{distill} the task-relevant logic and reasoning contained in $\mathcal{D}$, so that it can interpolate and generalize across tasks and unseen scenarios.

\subsection{Distilling reasoning with RNNs}\label{subsec:reasoning}

Recent work has established a theoretical connection between automata and a second-order RNNs to prove that a network with sufficient capacity can represent any automaton~\cite{li2024connecting}.
This connection can also be applied
to classical recurrent models like gated recurrent units, long short-term memory models or attention-based networks \cite{zhan_latmos:_2025}.
Consequently, we can train a sufficiently expressive recurrent model, in a supervised manner, to represent 
an automaton from $\mathcal{D}$---thus building a compact translation from a natural-language specification into an algorithm. 
This key insight underpins the model described in this section.

Given the dataset $\mathcal{D}$, we train the parameters $\psi$ of the RNN that operates according to 
\begin{equation}
    x_{k} = f_{\psi}\big(x_{k-1}, p^l_{i, k} \; \big | \, \mathcal{E}^l_i),
\end{equation}
where \( x_{k}\) is the internal state of the model at time $k$.
The goal of training $f_{\psi}$ is to encode 
the sub-tasks and transition logic of all automata $\{\mathcal{A}_i\}_{i \in \mathcal{T}}$ 
contained in $\mathcal{D}$.
By explicitly conditioning on $\mathcal{E}^l_i$ at every step, we ensure that incoming events and state transitions are learned in the context of $\mathcal{E}^l_i$.
This also ensures that latent states corresponding to different tasks remain separated in the latent space: 
even if two sequences from distinct tasks pass through identical event patterns, the task embedding $\mathcal{E}^l_i$ shifts their trajectories in the latent space so that $x_{k}$ for one task does not
overlap with \( x^{\prime}_{k} \) from another.

To learn $\psi$, a decoder projects $x_{k}$ into a predicted one-hot state representation $\hat{h}^{l}_{i,k} = d_{\phi}\left(x_{k}\right)$,
which is trained to approximate the ground-truth label $h^{l}_{i,k}$. 
The decoder $d_{\phi}$ is implemented as a single-layer MLP, deliberately kept shallow to ensure that all temporal reasoning and semantic grounding remains encoded in the latent of $f_{\theta}$. 
Finally, the loss $\mathcal{L}({h}^{l}_{i,k}, \hat{h}^{l}_{i,k})$ is a multi-class cross entropy loss
that penalizes incorrect state trajectories.

\subsection{Decentralized collaboration with GNNs}\label{subsec:collaboration}

Now that we have distilled the reasoning capabilities of LMMs into a single light-weight model, we must find policies that understand the latent representation of the automaton $x_k$ and interpret it to generate low-level control actions that solve the sub-task dictated by $x_k$. In our problem, we assume a Dec-POMDPs setting where we only have access to a reward signal.
Therefore, to solve Problem \ref{prob:prob} we propose a policy $\pi$ that is a composition of the RNN $f_{\psi}$ trained in Sec. \ref{subsec:reasoning} and a language-conditioned multi-task policy $g_{\vartheta}$, $\pi \equiv f_{\psi} \circ g_{\vartheta}$.
We learn $g_{\vartheta}$ using reinforcement learning.
Specifically, at time-step \(k\), the action of robot $r$ is
\begin{equation}\label{eq:control}
    a_{r,k} \sim g_{\vartheta}\big(a_{r, k} \,\big|\, \{o_{r^{\prime}, k}\}_{r^{\prime}\in N_{r,k}}, \; x_{r,k}\big),
\end{equation}
where
\( x_{r,k}\) is the RNN's latent state encoding the current sub-task,
and \(\{o_{r^{\prime}, k}\}_{r^{\prime}\in N_{r,k}}\) 
is an observation vector from robot $r$ and its neighbors at time $k$.
We denote the latent state with a subscript $r$ because, for a decentralized implementation, each robot carries its own local copy of the RNN.
This not only enables decentralized execution, but also lets robots
use local information to step through their automata asynchronously when needed.
On the other hand, the policy only depends on the local observation of robot $r$ and information received from neighbors.
Therefore, we parameterize $g_{\vartheta}$ as a GNN that naturally handles time-varying communication graphs.

In learning $\vartheta$, we take advantage of the dataset, where the RL scenarios are automatically paired with sentence embeddings.
This lets us apply standard multi-agent reinforcement learning algorithms like Multi-Agent Proximity Policy Optimization \cite{yu2022surprising} to train the multi-task policy, provided that every episode randomizes the sub-tasks and the reward signal $R_{i,h}$ accordingly.
Hence, unlike existing multi-task policies that operate over entire episodes, we train $g_{\vartheta}$ to act over short temporal segments corresponding to individual sub-tasks.
At execution time, an appropriate sub-task policy is selected by conditioning on the latent state of the RNN, which encodes the sub-task in the task automaton.
With a sufficiently expressive model $g_{\vartheta}$ we can capture the solution of all sub-tasks; by restricting training to individual sub-tasks, the computational load during training is reduced.
High-level decision-making is delegated entirely to the RNN, allowing the multi-task policy to focus on low-level execution in the context of its current sub-task. 

Lastly, $o_r$ contains information from sub-task related measurements (e.g., position of obstacles) and event information related to the atomic propositions of the automata (e.g., a switch turns on).
The latter can be shared among neighbors to improve the reactivity of the local RNN at each robot and the efficiency of the team overall.
We therefore modify the input to the RNN from the observed $p^l_{i,k}$ to an aggregated $\Bar{p}^l_{i,k}$ that includes information from neighbors as well.
The vector $\Bar{p}^l_{i,k}$ is obtained from a general aggregation operation $\doubleoplus$ (see Fig. \ref{fig:overview} (middle and bottom)) that can be implemented in multiple ways.
In the most general case, $\doubleoplus$ can be learned, for instance, as another GNN.
For some use-cases, it can be reduced to simple boolean logic such as \texttt{AND} or \texttt{XOR}; or it can even be extracted automatically from the language command, although we leave this option as future work. 

%%%%%%%%%%%%%%%%           
%% EVALUATION %%
%%%%%%%%%%%%%%%%

\section{Evaluations}\label{sec:evaluation}

Our pipeline accepts as input a task as a natural-language command, and leads a robot team to execute it.
We encode numerous task automata into a single RNN that represents their logical decompositions,
and train the decentralized policy conditioned on RNN states. 
The combination of the two makes for a powerful recipe---we bridge the gap between language-driven \textit{reasoning} and its real-time distributed \textit{implementation} as feedback control.
Thus, we now evaluate our framework for
\textit{(i)} its correctness in representing the various task automata, and,
\textit{(ii)} its ability to complete a given multi-robot task, even when presented with disruptions, potentially long/repeated sub-task sequences, and real-world noise.
We use the \textit{same} RNN to encode all the tasks presented in this section\footnote{All parameters and tasks to reproduce the evaluations are on our \href{https://sites.google.com/view/prompting-teams}{website}.}.

\subsection{Automata Representation}\label{subsec:eval_e2}
\begin{figure}
    \DeclareRobustCommand{\automtermstate}{
        \tikz[baseline=-0.75ex]{
            \node[shape=circle,fill,minimum size=1.0ex, inner sep=.5pt] () {};
            \node[shape=circle,draw,minimum size=1.5ex, inner sep=.5pt] () {};
        }
    }
    \centering
    \includegraphics[width=\linewidth]{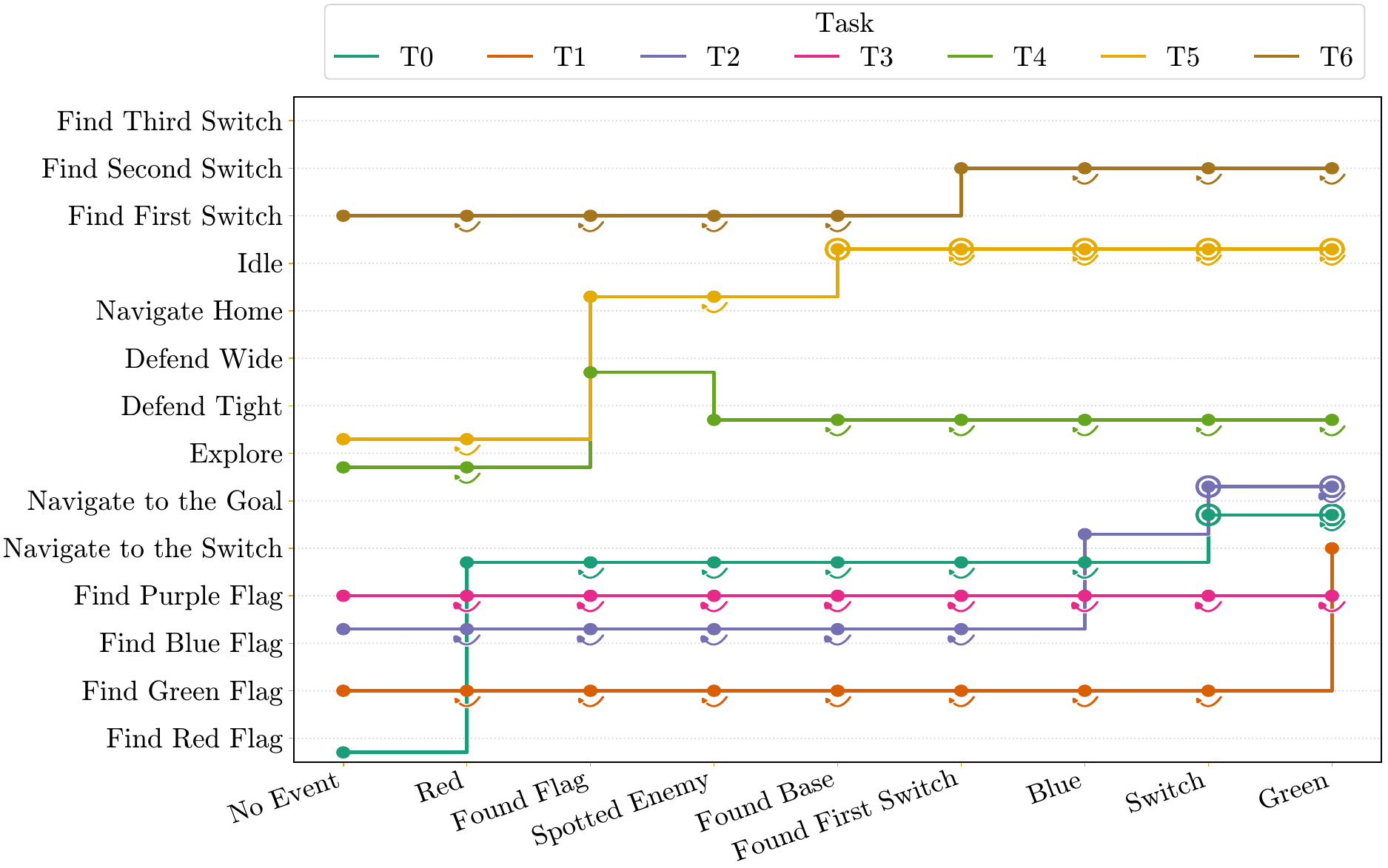}
    \caption{
        Rollouts of the same RNN on seven tasks (\textbf{T}\{0..6\}) over the available sub-tasks (y-axis) with relevant/irrelevant events presented to it (x-axis).
        A \automtermstate denotes absorbing states.
        Training and test achieve $100\%$ accuracy: the model rejects events that belong to a potentially different automaton.
        For instance, \textcolor[HTML]{D95F02}{\textbf{T1}} requires the team to \textit{``Search for the green flag, then the switch, and navigate to the goal''}, and events such as \textit{``Spotted Enemy''} have no impact.
        The same event, ocurring in \textcolor[HTML]{66A61E}{\textbf{T4}}, \textit{``Locate the mission flag and defend the position; adapt your defense.''}, causes the team to change its formation from \textit{`wide'} to \textit{`tight'}.
        Other tasks are listed on our website.
    }
    \label{fig:e2-autom-rollouts}
\end{figure}
{Since the RNN encodes numerous automata, we first want}
that given a language input, $\mathcal{W}_i$, the model to be initialized into the correct state, $h_{i,0}$, of automata $\mathcal{A}_i$.
Then, subsequent state transitions must follow the logic of the task, and never produce a wrong state $h_{i^{\prime},k}$ from a different task $i \neq i^{\prime}$ even when presented with events from 
$i^{\prime}$. {The RNN is a Gated Recurrent Unit with a 1024 hidden layer and a projected language embedding with a $[512, 512, 64]$ MLP.}

\Cref{fig:e2-autom-rollouts} shows evaluation sequences
for seven different tasks, $\mathrm{\mathbf{T}\{0..6\}}$,
as the model steps through the automata pool and the sub-tasks on the vertical axis when presented with the sequence of (ir-)relevant events on the horizontal axis.
For instance, in \textcolor[HTML]{7570B3}{\textbf{T2}}:
\textit{``Find the blue flag, spot the switch, and head for the target'',} the first sub-task is \textit{`Find blue flag'}, and the RNN remains in this state, {rejecting irrelevant events until the observation \textit{`Blue'}.}
It then transitions to \textit{`Navigate to the Switch'}, and similarly, remains in this state until \textit{`Switch'}.
We observe that for each task, the initialization of the correct sub-task follows the task's semantics, i.e., given \textit{`No Event'}, the first sub-task always corresponds to the relevant one according to $\mathcal{W}_i$. This is of key importance because it means that the RNN does not {initialize an incorrect automaton.}

We also observe that the rollout for each task follows a logically correct and semantically relevant sequence of sub-tasks.
The RNN transitions occur only for relevant events, and {all other irrelevant events from other tasks are rejected}.
For instance, \textit{`Found Base'} and \textit{`Spotted Enemy'} only produce self-loops in \textcolor[HTML]{7570B3}{\textbf{T2}}.
Indeed, our supervised train and test accuracy is a \SI{100}{\percent}, observed on 500 sequences per task with an 80-20 split. 
Note that the sequence of events (horizontal axis) has no particular order, and can continue indefinitely.
We list the full set of tasks and events on the \href{https://sites.google.com/view/prompting-teams}{website}.
The representation capacity of the RNN indicates that it is possible to learn hundreds or even thousands of tasks in a single model \cite{zhan_latmos:_2025, butoitraining}.
A procedural method that automatically ``imagines'' tasks for robot teams could potentially be used to extract numerous automata and distill them all into our RNN.

\subsection{Ablation: Need for Recurrence and Language}\label{subsec:eval_e0}
\begin{figure*}
    \centering
    \begin{minipage}[c]{0.32\linewidth}
    \begin{tikzpicture}
        \draw (0, 0) node[inner sep=0]{
            \tcbox[size=fbox]{
                \includegraphics[width=0.47\linewidth]
                    {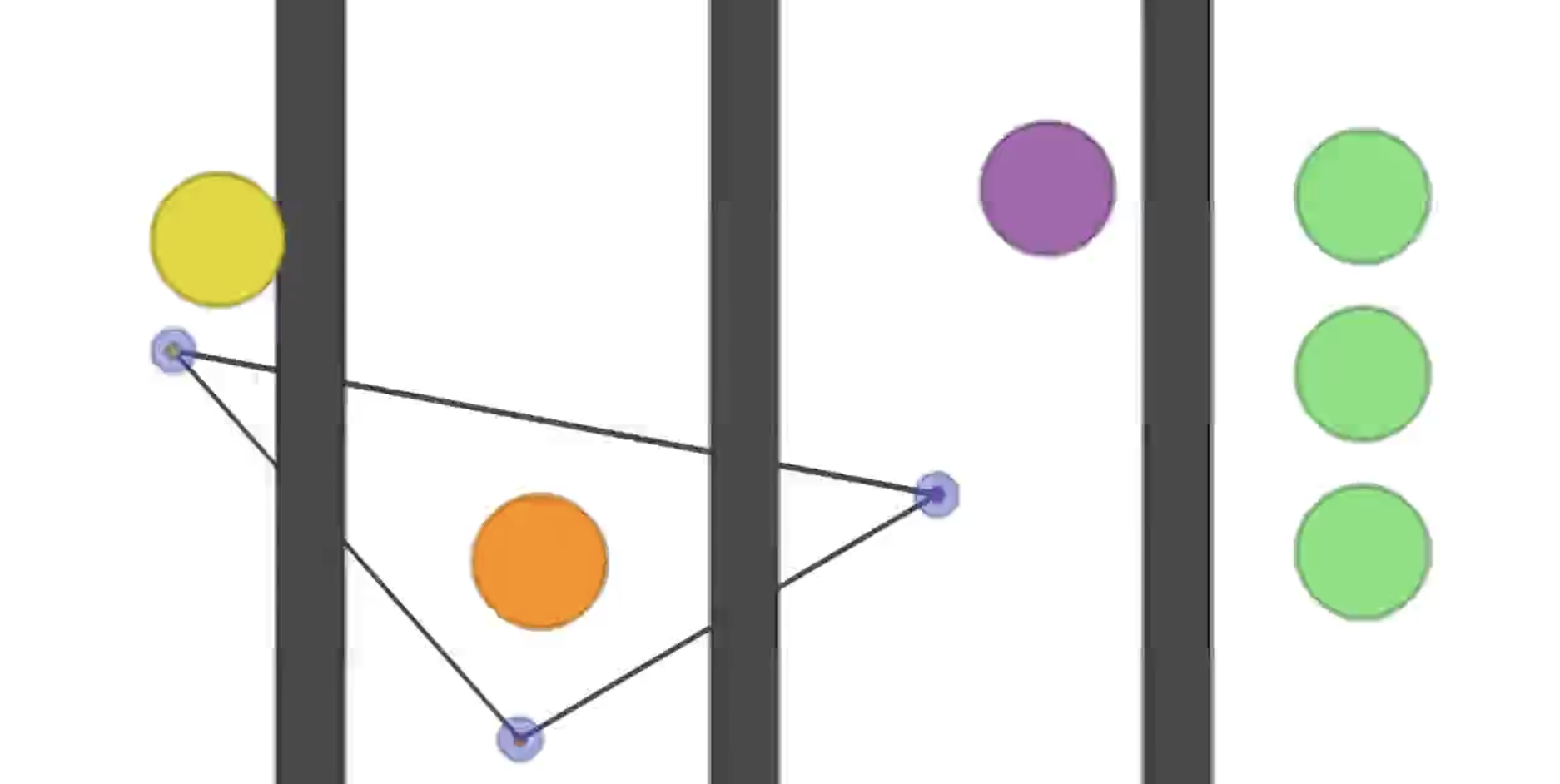}
            }
        };
        \draw (0.8,-0.55) node[fill=white,opacity=.75]{
            \textbf{\footnotesize t=0s}
        };
    \end{tikzpicture}%
    \begin{tikzpicture}
        \draw (0, 0) node[inner sep=0]{
            \tcbox[size=fbox]{
                \includegraphics[width=0.47\linewidth]
                    {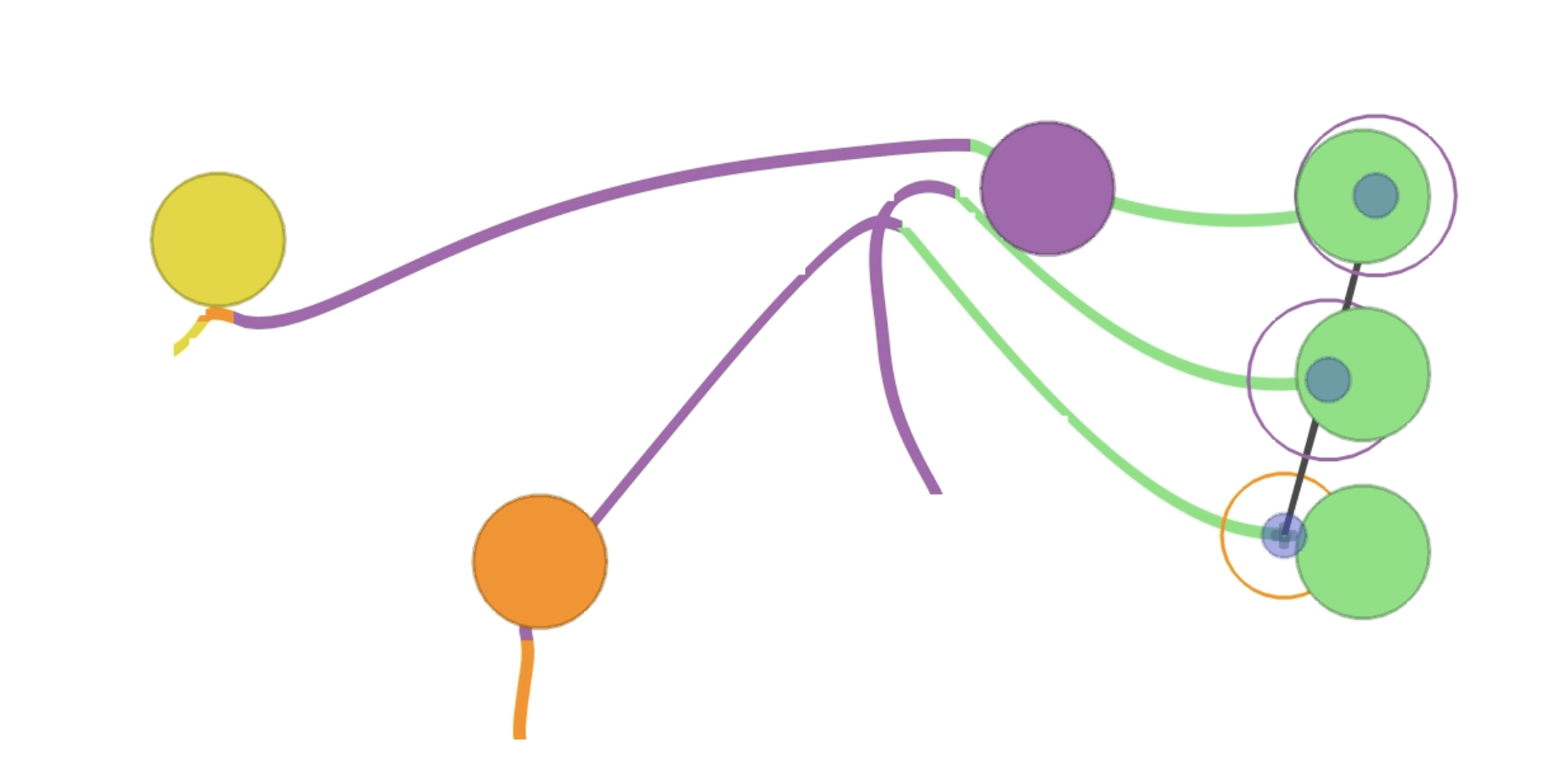}
            }
        };
        \draw (0.8,-0.55) node[fill=white,opacity=.75]{
            \textbf{\footnotesize t=13s}
        };
    \end{tikzpicture}\\%
    \begin{tikzpicture}
        \draw (0, 0) node[inner sep=0]{
            \tcbox[size=fbox]{
                \includegraphics[width=0.47\linewidth]
                    {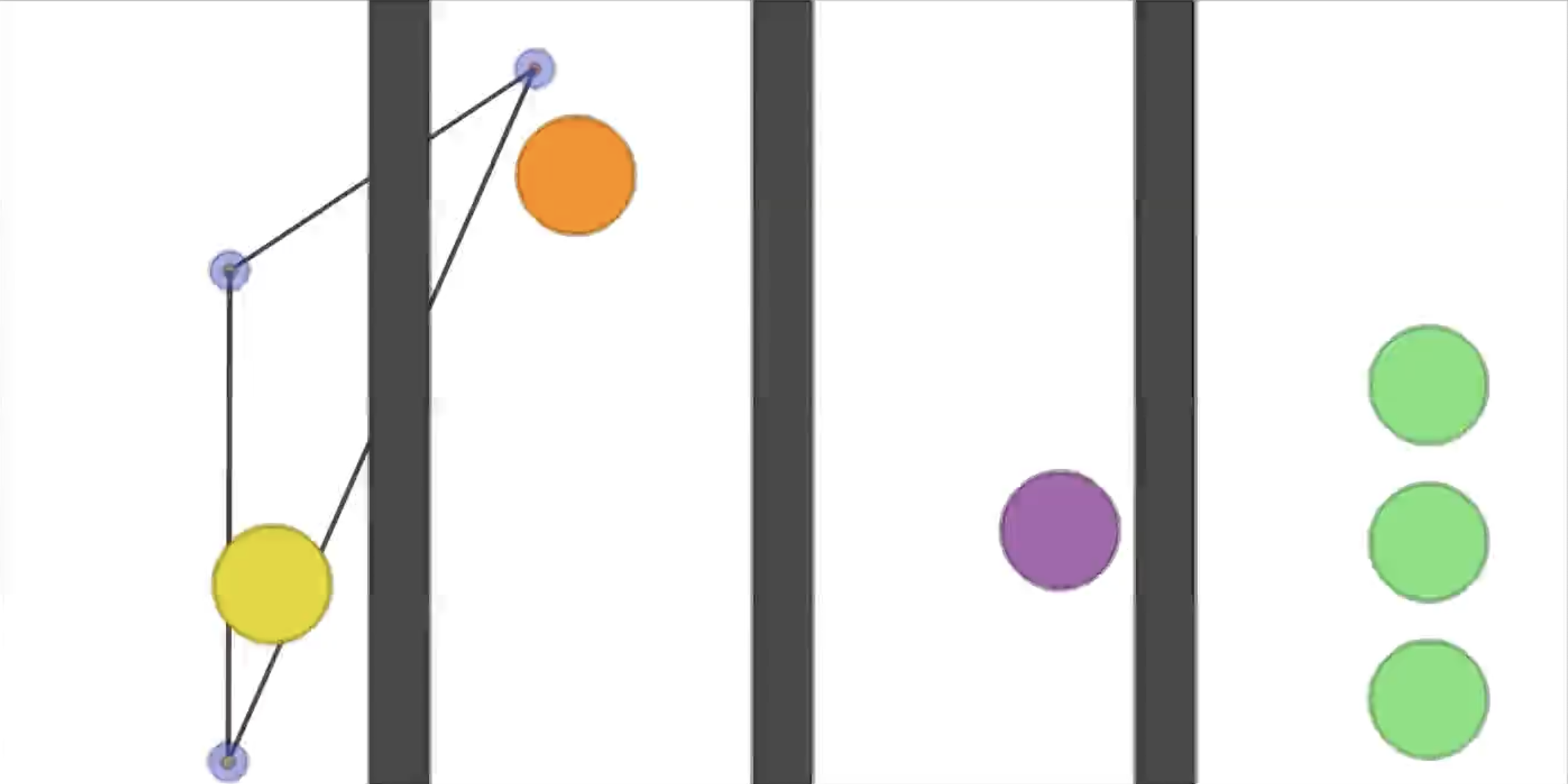}
            }
        };
        \draw (0.8,-0.55) node[fill=white,opacity=.75]{
            \textbf{\footnotesize t=0s}
        };
    \end{tikzpicture}%
    %\hfill
    \begin{tikzpicture}
        \draw (0, 0) node[inner sep=0]{
            \tcbox[size=fbox]{
                \includegraphics[width=0.47\linewidth]
                    {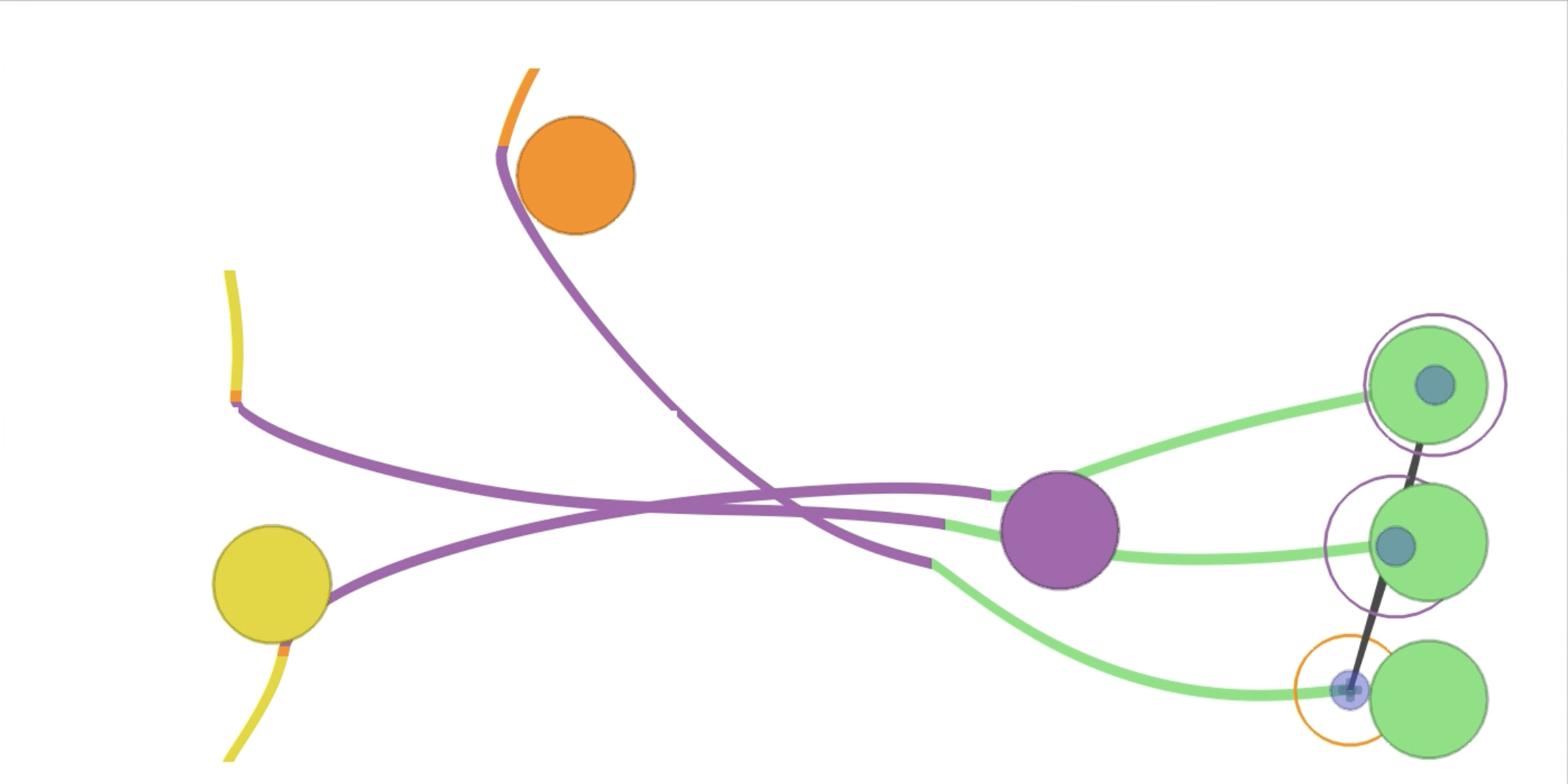}
            }
        };
        \draw (0.8,-0.55) node[fill=white,opacity=.75]{
            \textbf{\footnotesize t=13s}
        };
    \end{tikzpicture}\\%
    \begin{tikzpicture}
        \draw (0, 0) node[inner sep=0]{
            \tcbox[size=fbox]{
                \includegraphics[width=0.47\linewidth]
                    {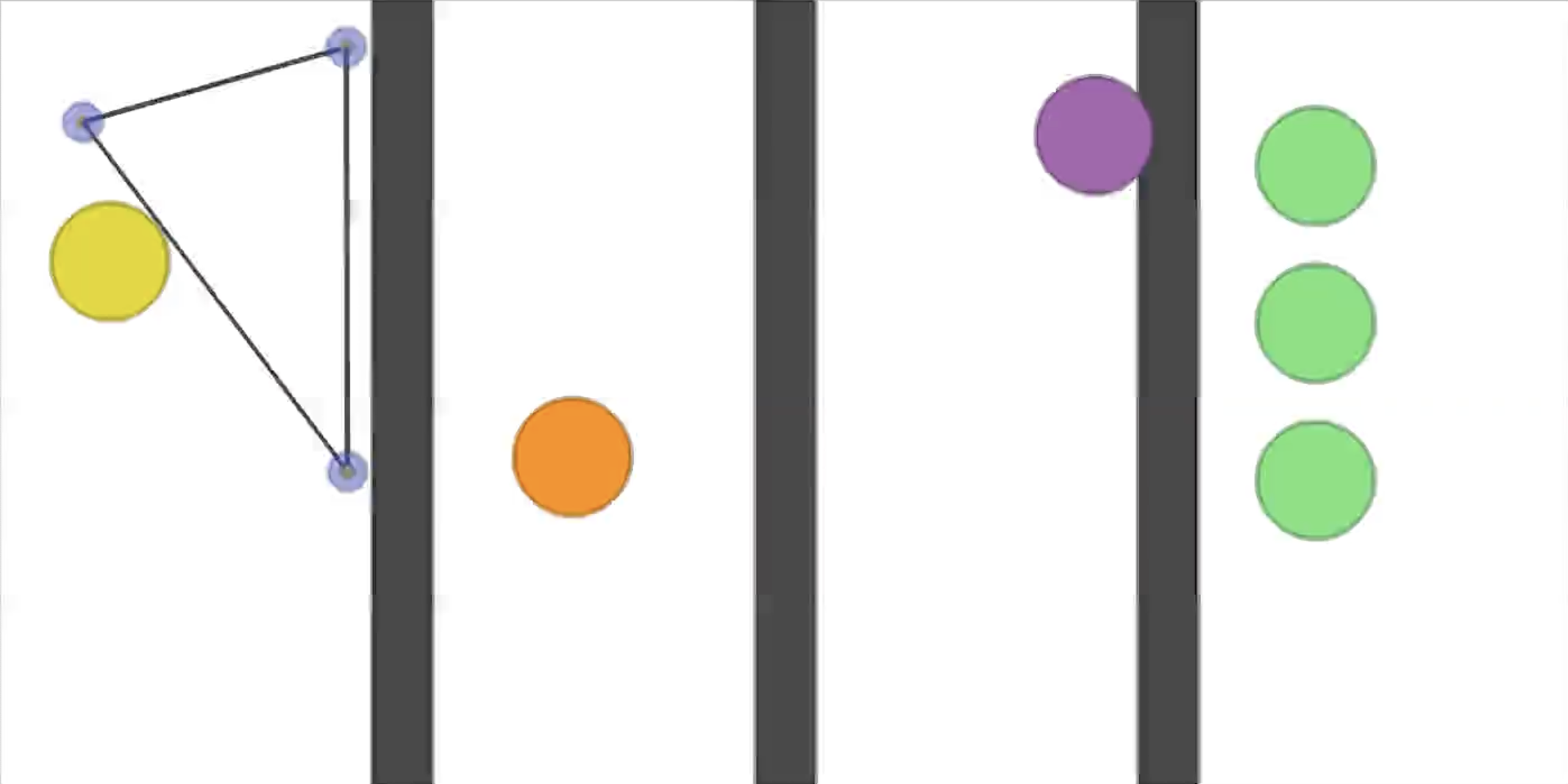}
            }
        };
        \draw (0.8,-0.55) node[fill=white,opacity=.75]{
            \textbf{\footnotesize t=0s}
        };
    \end{tikzpicture}%
    %\hfill
    \begin{tikzpicture}
        \draw (0, 0) node[inner sep=0]{
            \tcbox[size=fbox]{
                \includegraphics[width=0.47\linewidth]
                    {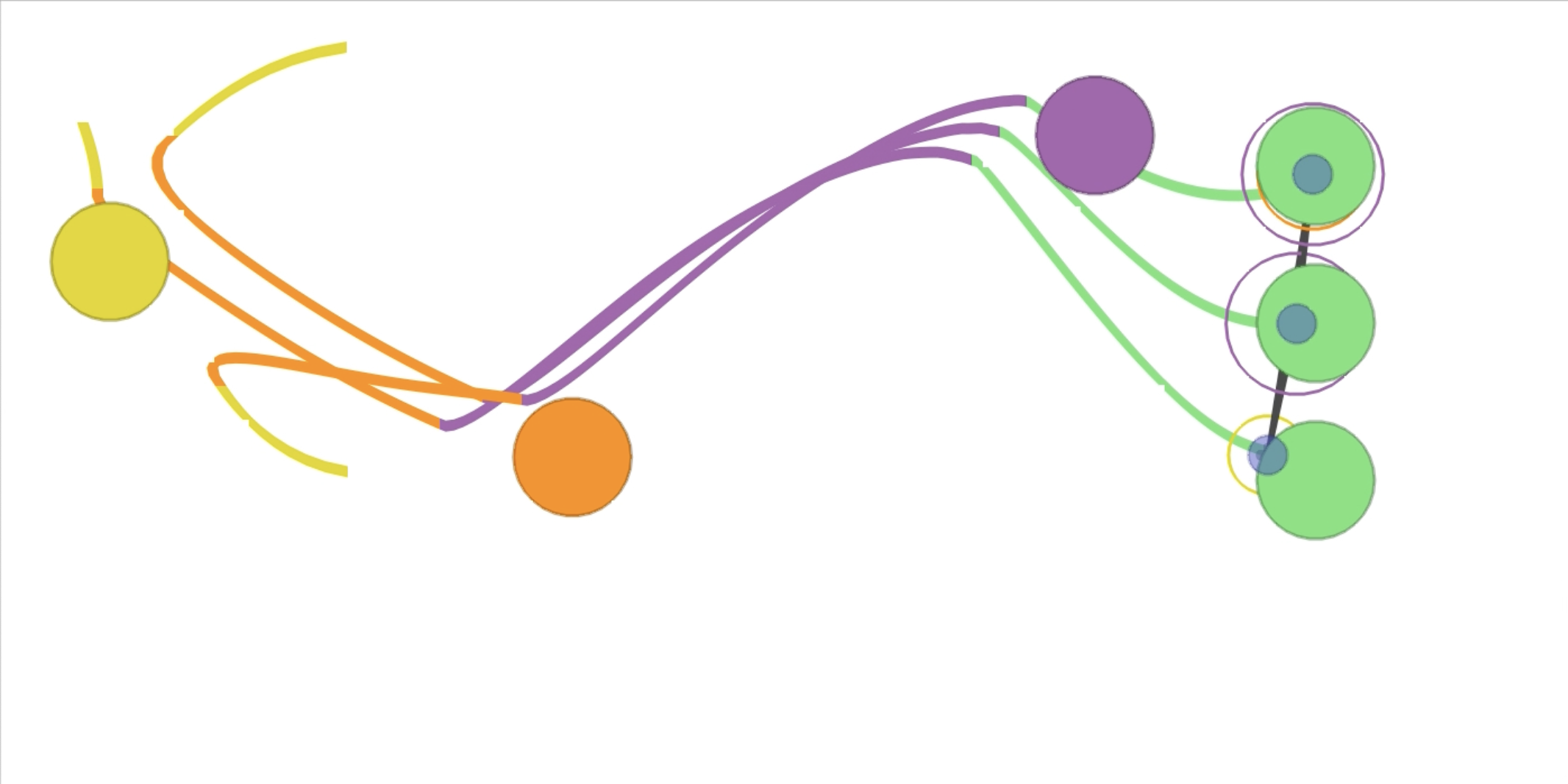}
            }
        };
        \draw (0.8,-0.55) node[fill=white,opacity=.75]{
            \textbf{\footnotesize t=13s}
        };
    \end{tikzpicture}%
    \end{minipage}%
    \hfill\hspace{0.00\linewidth}%
    \begin{minipage}[c]{0.66\linewidth}
        \includegraphics[width=0.49\linewidth,trim={5pt, 5pt, 2pt, 2pt},clip]
            {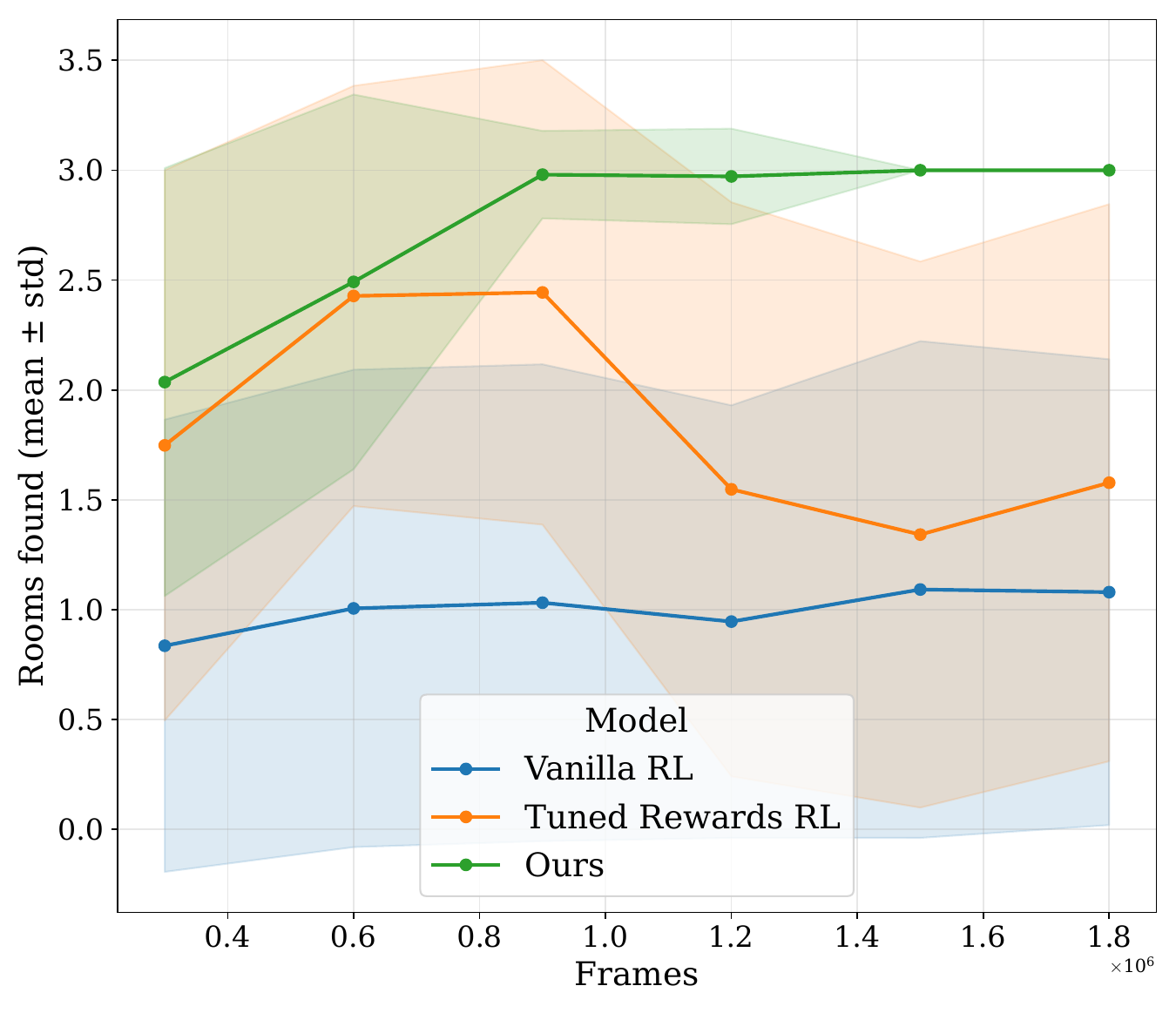}
        \includegraphics[width=0.5\linewidth,trim={20pt, 20pt, 2pt, 2pt},clip]
                     {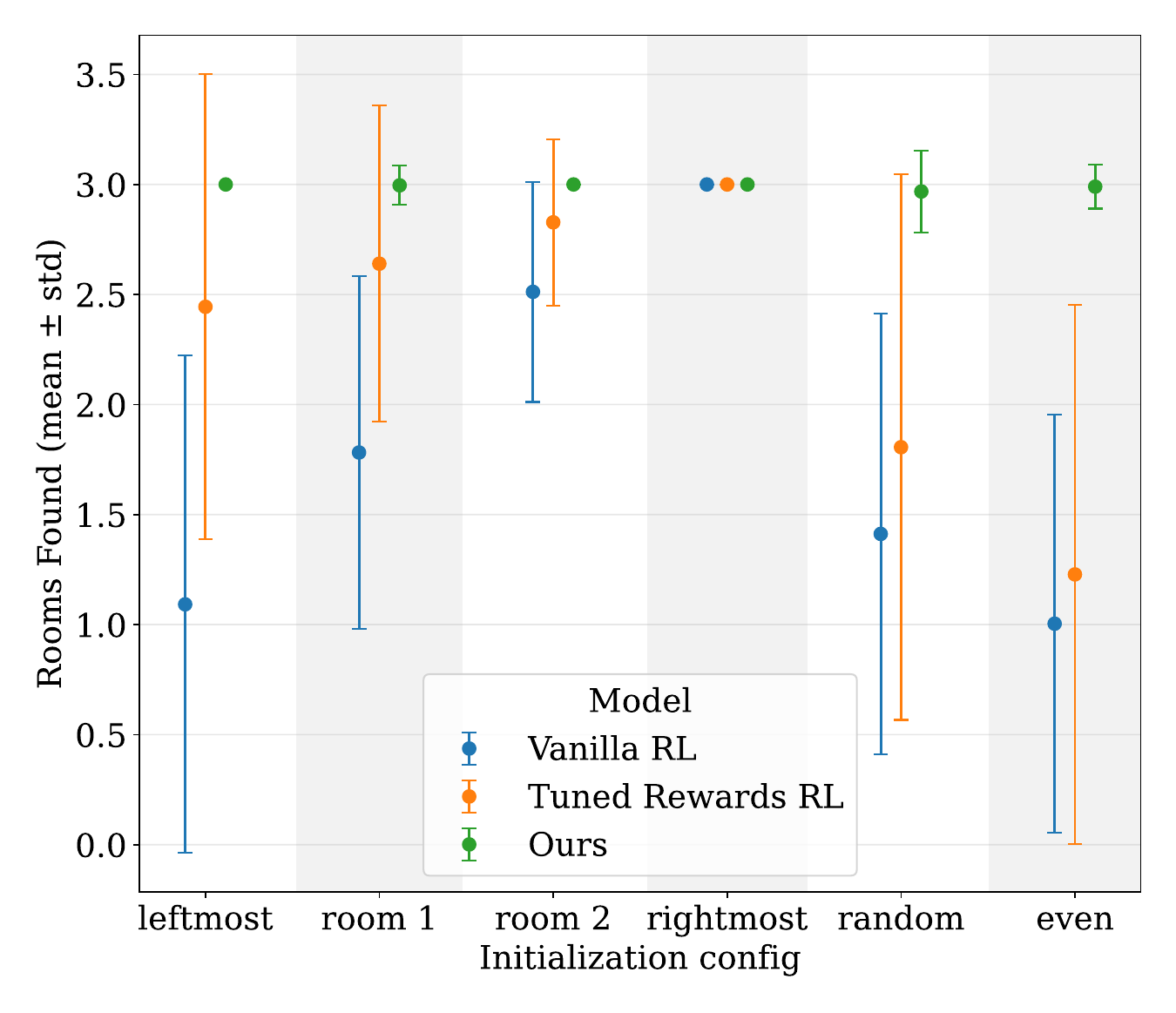}
    \end{minipage}
    \caption{Analysis of the impact of reasoning and language in a four-room scenario. Three robots are placed in any of the three leftmost rooms, which are divided by solid black walls. Each room has a differently colored switch (\textcolor{yellow}{\textbullet} \textcolor{Apricot}{\textbullet} \textcolor{violet}{\textbullet}), and the robots need to hit them in order to open a passage and navigate to their goals (\textcolor{green}{\textbullet}). (left) Snapshots of different robot and RNN initializations, showing that our solution successfully accomplishes the task irrespective of the configuration of robots, switches and goals. Traces are colored according to the sub-tasks deducted by their local RNNs. (middle) Training curves for the three benchmarked methods: a vanilla RL policy, a reward-tuning RL policy, and our policy; success is measured as the average number of rooms found by the team of robots in 500 episodes. Our policy consistently outperforms the other methods by achieving perfect task completion at the end of the training. (right) After training, we evaluate all six possible initial configurations with the three methods.
    }
    \label{fig:e0_multiroom_evals}
\end{figure*}

{Since the RNN can correctly step through complex, multi-step task automata, we now focus on evaluating its necessity in task completion, especially when using language inputs.}
We use the scenario shown in \Cref{fig:e0_multiroom_evals}(left), where a team of three robots is initialized at random locations in four rooms.
Each of the first three rooms on the left has a switch that a robot can hit to gain passage into an adjacent room on the right, to ultimately reach its assigned goal (in the rightmost room).
The team receives a natural language description of the overall task,%
for instance, \textit{``Robots, unlock the first, second, and third switches in sequence, then advance to the objective room and reach the target.''}
Any individual robot may be further initialized with a sub-task as: \textit{``Head to the second trigger in the second room from the left.''}

To ablate recurrence, we train a `Vanilla RL' baseline that only rewards the team when the robots navigate correctly to their goals, and has no reasoning or an explicit model for the sequence of sub-tasks.
We separately ablate language and automata by training a `Tuned Rewards RL' baseline with a hand-crafted multi-stage reward that {explicitly decomposes the task.}
This causes overfitting to a sequence of sub-tasks: the team cannot exploit semantic hints from language and will require a new reward structure for each new configuration. {All methods use a Graph-Attention GNN \cite{velivckovic2018graph} with a single head of dimensions $[64,256,256,4]$. }

In \Cref{fig:e0_multiroom_evals} (middle), we see a comparison (averaged over $500$ runs) of our method against these two candidate baselines.
`Vanilla RL' performs well only in hitting the very first switch, after which, the successive rooms are never unlocked.
The Tuned-Rewards RL method performs significantly better, since we already decompose rewards according to the stages in the task.
This decomposition is manual, and thus laborious, error-prone, and infeasible to scale and apply to different scenarios.
Our method sidesteps this problem by {encoding the appropriate sub-tasks in the RNN (rooms, switches)}, and delegating the sub-task switching to it.

Our method also enables robots to be initialized non-uniformly (same/different rooms), both physically as well as in their initial RNN states through natural language, as 
seen in the three samples in \Cref{fig:e0_multiroom_evals}(left).
We show quantitative analysis in \Cref{fig:e0_multiroom_evals}(right) by initializing robots in all six unique configurations for each method, {trained for \num{1e6} frames} and evaluated over $500$ runs.
Clearly, when the team is initialized in the rightmost room, all methods perform equally well ($=\SI{100}{\percent}$), since the only sub-task then is to navigate to their respective goals.
Similarly, initialization in `room-2' is only slightly worse since it adds only one additional sub-task.
For all other initializations, only the language-driven RNN method succeeds in completing the entire task reliably---especially in the hardest configuration where the robots are spread evenly (one in each room).
Although the baseline, language-free RL methods eventually learn under predictable initializations, their catastrophic failures under ``random'' and ``even'' initializations persist, as these require asymmetric behaviors that are considerably harder to learn.

{We note that} our method does exhibit some susceptibility in the `random' and `even' configurations.
We attribute this to {highly divergent language initializations:}
a sufficiently {meaningless} expression can produce an incorrect mapping into an automaton's state.
As an instance from our \href{https://sites.google.com/view/prompting-teams}{website}, \textit{``Exploration drones, activate the three room switches from left to right; then move into room four and finalize at the goal''} with a sub-task \textit{``Units, make your way smoothly to the second release plate in the room after the left-most bay near the middle divider gate; signal ready when the ring glows''}.

\subsection{{Comparison with Baselines in an Unstructured Task}}
\label{subsec:search-rescue}

{We now test and benchmark the robustness of our method in handling more complex automata with parallel and heterogeneous sub-tasks for the team, partial language specification, and event-space disruptions.
As baselines, we use
\textit{(i)} an online \textsc{llm} planner (Llama~3.3~70B  \cite{grattafiori2024llama}) to generate per-agent actions when called every 20 environment steps and provided with the language instruction, world state, and recent events;
\textit{(ii)} the GRU variant of \textsc{latmos}~\cite{zhan_latmos:_2025}, and,
\textit{(iii)} a distilled Small Language Model \textsc{(slm)} based on the encoder of \texttt{flan-t5-small}~\cite{flan_t5}.
The last two methods are fine-tuned in the same way our method trains the RNN, and matched to a low-level decentralized policy using MAPPO as ours.}

{We design a task for a heterogeneous team of four robots, two scouts and two carriers, that need to discover two targets in a 2D area, and transport them to their corresponding (assigned) medical bays.
Only scouts can discover targets, and a target can be transported only when two carriers lift together.
The resultant automaton
necessitates parallelism, and exhibits several branches.}
{We 
encode four different DFA families corresponding to four prompt groups:
\texttt{either} target may be rescued first,
target-\texttt{1-first}, 
target-\texttt{2-first}, and, 
\texttt{mutex} prohibiting scouts to search for target 2 before target 1 is rescued. 
Prompts can be as abstract as: \textit{``The orange casualty is more urgent.''}
}

{Table~\ref{tab:search-rescue-main} reports success rates at two episode budgets.
All trained methods fully solve the task when given the $800$-step budget, however, differences emerge with a tighter $250$-step budget.
Our $0.16$M-parameter model remains competitive against the distilled \textsc{slm} and \textsc{latmos} (with $168\times$ and $10\times$ more parameters, respectively). 
For the strict constraint, the two language-conditioned methods complete $60\%$ of the tasks, while \textsc{latmos}, which has no language input by construction, drops to $40\%$.
The gap is prominent here, since the constraint cannot be inferred from observations alone. 
The online \textsc{llm} planning fails this cooperative-carry task entirely:
the model issues reasonable discovery commands, but cannot maintain synchronization between the carriers required for transport; the failure mode is coordination, not time-budget. }

{\newcolumntype{C}{>{\centering}p{6ex}}
\begin{table}[t]
\centering
\caption{{Success rates on the search-and-rescue task measured at 800 and 250 steps, $20$ episodes per variant. Encoder parameter counts for inference.}}
\label{tab:search-rescue-main}
\setlength{\aboverulesep}{0pt}
\setlength{\belowrulesep}{1pt}
\renewcommand{\arraystretch}{1.0}
\begin{tabular}{@{}l|c|CCCC@{}}
\toprule
       &         \multicolumn{1}{c|}{$800$-steps} & \multicolumn{4}{c}{$250$-steps} \\
Method (Encoder) &  & \texttt{either} & \texttt{1-first} & \texttt{2-first} & \texttt{mutex}\tabularnewline
\midrule
Ours            \hfill ($0.16$M)   & ${100\%}$ & $65\%$  & $30\%$   & $45\%$  & ${60\%}$ \tabularnewline
\textsc{slm}    \hfill ($26.8$M)   & ${100\%}$ & $60\%$  & ${45\%}$ & $50\%$  & $60\%$   \tabularnewline
\textsc{latmos} \hfill ($1.5$M)    & ${100\%}$ & $55\%$  & $35\%$   & $50\%$  & $40\%$   \tabularnewline
\textsc{llm}    \hfill ($70$B)     & $0\%$     & $0\%$   & $0\%$    & $0\%$   & $0\%$    \tabularnewline
\bottomrule
\end{tabular}
\end{table}
}

\begin{table}[t]
\centering
\caption{{Success across all variants (10 episodes per variant, 250-step budget) under held-out paraphrases and $20\%$ event-drop noise.}}
\label{tab:search-rescue-robust}
\setlength{\aboverulesep}{0pt}
\setlength{\belowrulesep}{1pt}
\renewcommand{\arraystretch}{1.0}
\begin{tabular}{lccc}
\toprule
Method & Unperturbed & Held-out paraphrase & $20\%$ event-drop \\
\midrule
Ours             & $50.0\%$ & $50.0\%$ & $42.5\%$ \\
\textsc{slm}     & $53.8\%$ & $60.0\%$ & $52.5\%$ \\
\textsc{latmos}  & $45.0\%$ & $47.5\%$ & $40.0\%$ \\
\bottomrule
\end{tabular}
\end{table}

{We evaluate robustness against two sources of errors: unseen language variations, and random event disruptions.
We create 6 held-out paraphrases per DFA variant that are semantically equivalent to those in $\mathcal{D}$, and substitute their corresponding $\mathcal{E}$ randomly during rollout.
For emulating event disruptions, at each environment step, we drop any event bit set to $1$ with some probability $p$.
Both perturbations leave success rates within a few percentage points of the unperturbed baseline (Table~\ref{tab:search-rescue-robust}).
Notably, paraphrase substitution does \textit{not} cause systematic degradation, indicating that the RNN encoder understands task semantics. 
The event-noise result is consistent with the policy treating events as soft signals that can be re-discovered.}

\subsection{Long Sequences, Disruptions and Zero-shot Deployment}\label{subsec:real_exp}

\begin{figure}
    \centering
    \begin{tikzpicture}
        \draw (0, 0) node[anchor=north west,inner sep=0] (POS) {
            \includegraphics[width=\linewidth]
                {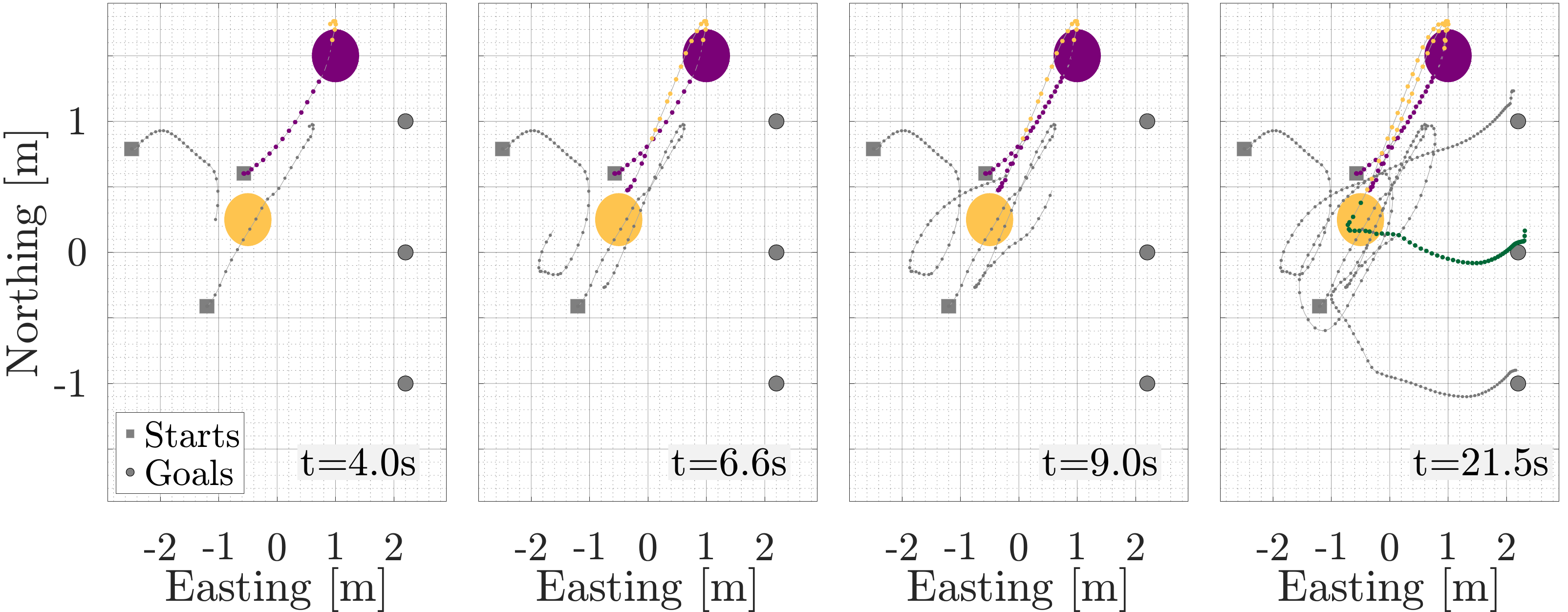}
        };
        \draw node[anchor=north west,inner sep=0,below=1ex of POS] (ERR) {
            \includegraphics[width=\linewidth]
                {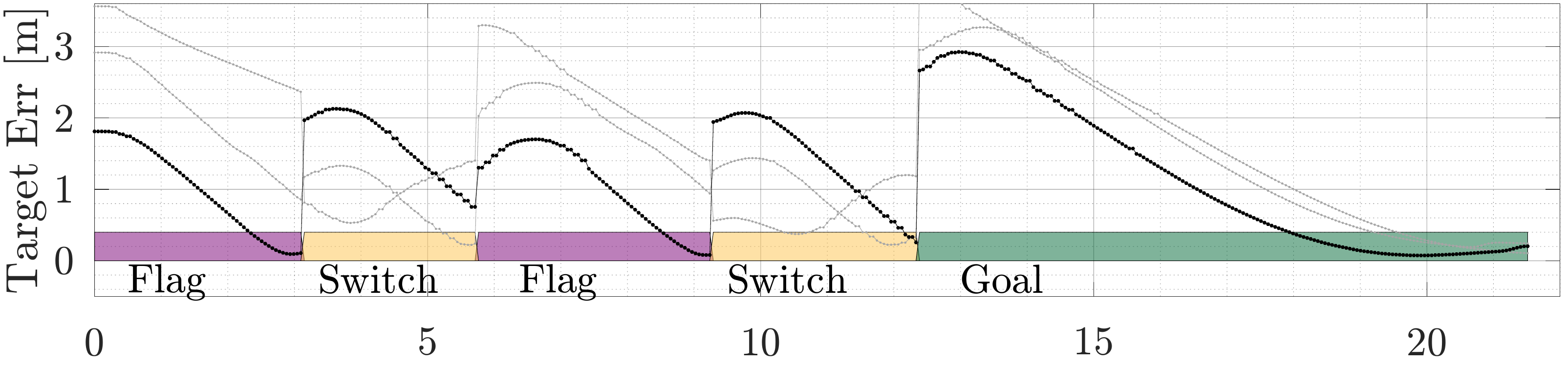}
        };
        \draw node[fill=white,anchor=north west,inner sep=0, below right=-1.6ex and -8ex of ERR] {
            \scriptsize Time $\mathrm{[s]}\rightarrow$%
        };
    \end{tikzpicture}
    \caption{Evolution of team trajectories from an instance of `retrieve-the-flag' scenario (\Cref{fig:hero}), with the team initialized at random locations and assigned fixed goals.
    The task requires them to ``\textit{Find a purple flag, and bring it to the switch}'', which then permits to cross to their goal locations.
    \textbf{(top)} Four snapshots of a top-down view at different times show the spatial evolution of the task.
    \textbf{(bottom)} A temporal view of the task shown as a function of robot distances to their current targets (Flag, Switch or Goal).
    The task faces a disruption (`Flag lost') at $\approx\SI{6}{s}$ before they reach the switch, causing the team to double back to the flag location a second time.}
    \label{fig:e3-realworld-evals}
\end{figure}
\begin{figure*}[t]
    \centering
    % -------- Row 1: real world --------
    \begin{subfigure}[t]{0.15\textwidth}
        \centering
        \includegraphics[width=\linewidth,trim={50px, 0, 66px, 0},clip]
            {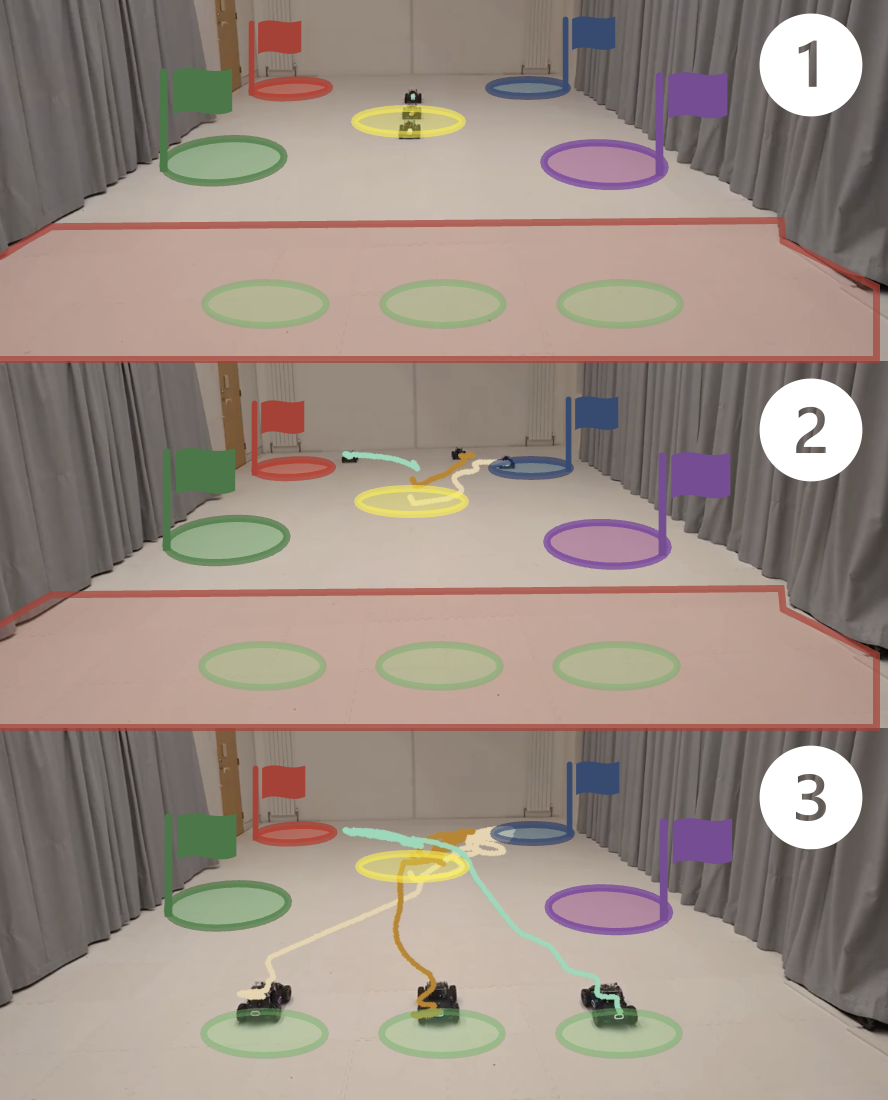}
        \caption{``Search for the blue flag, then the switch, then head to the goal room.''}
    \end{subfigure}\hfill
    \begin{subfigure}[t]{0.15\textwidth}
        \centering
        \includegraphics[width=\linewidth,trim={50px, 0, 66px, 0},clip]
            {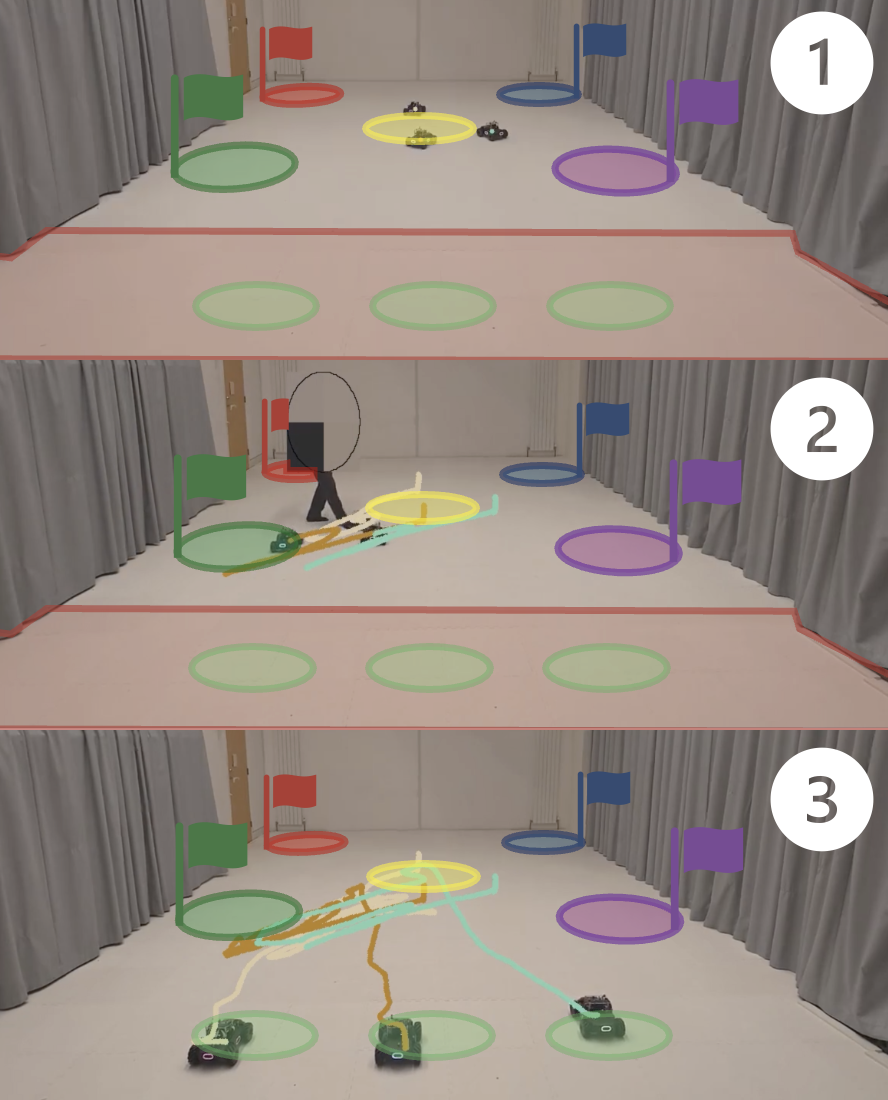}
        \caption{``First search for the green flag, then the actuator, then proceed to goal.''}
    \end{subfigure}\hfill
    \begin{subfigure}[t]{0.15\textwidth}
        \centering
        \includegraphics[width=\linewidth,trim={50px, 0, 66px, 0},clip]
            {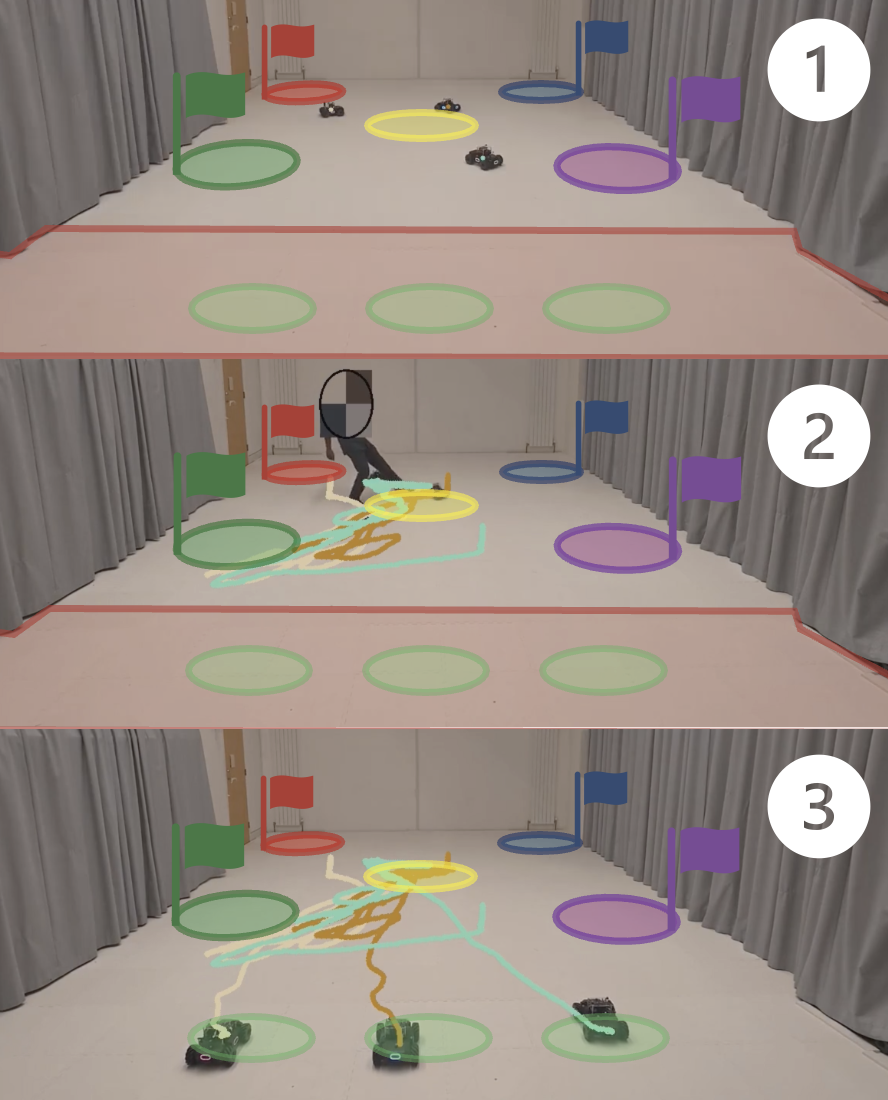}
        \caption{Same as (b), with major disruptions.}
    \end{subfigure}\hfill
    \begin{subfigure}[t]{0.156\textwidth}
        \centering
        \includegraphics[width=\linewidth]{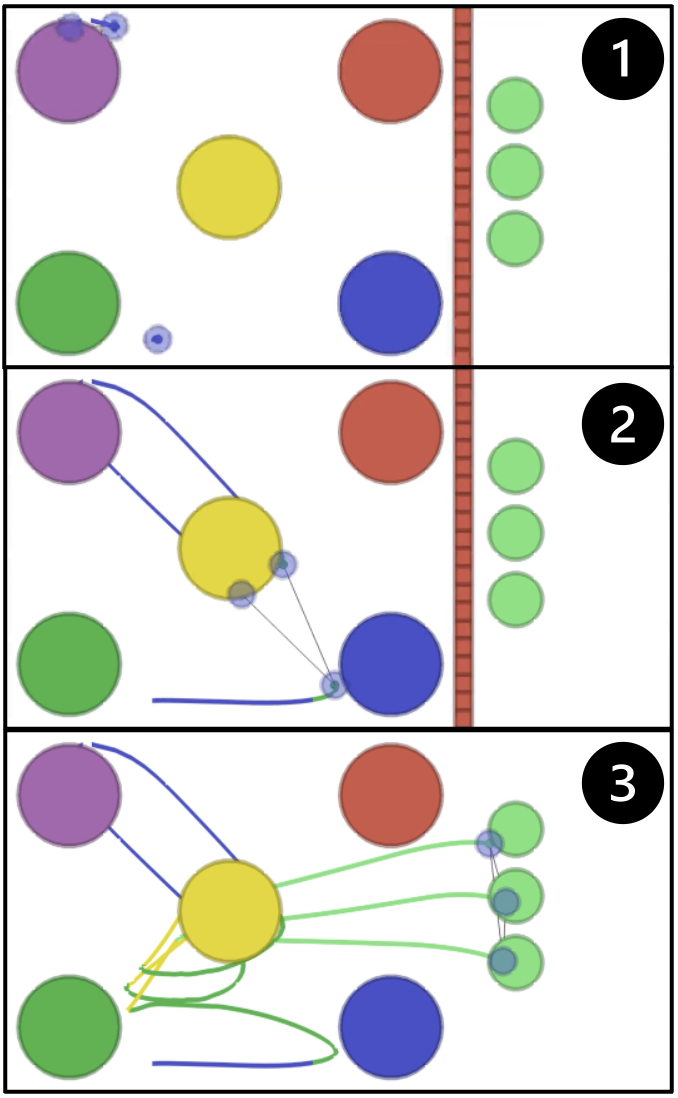}
        \caption{``Head for the blue flag, then the green flag, and finish at the target.''}
    \end{subfigure}\hfill
    \begin{subfigure}[t]{0.156\textwidth}
        \centering
        \includegraphics[width=\linewidth]{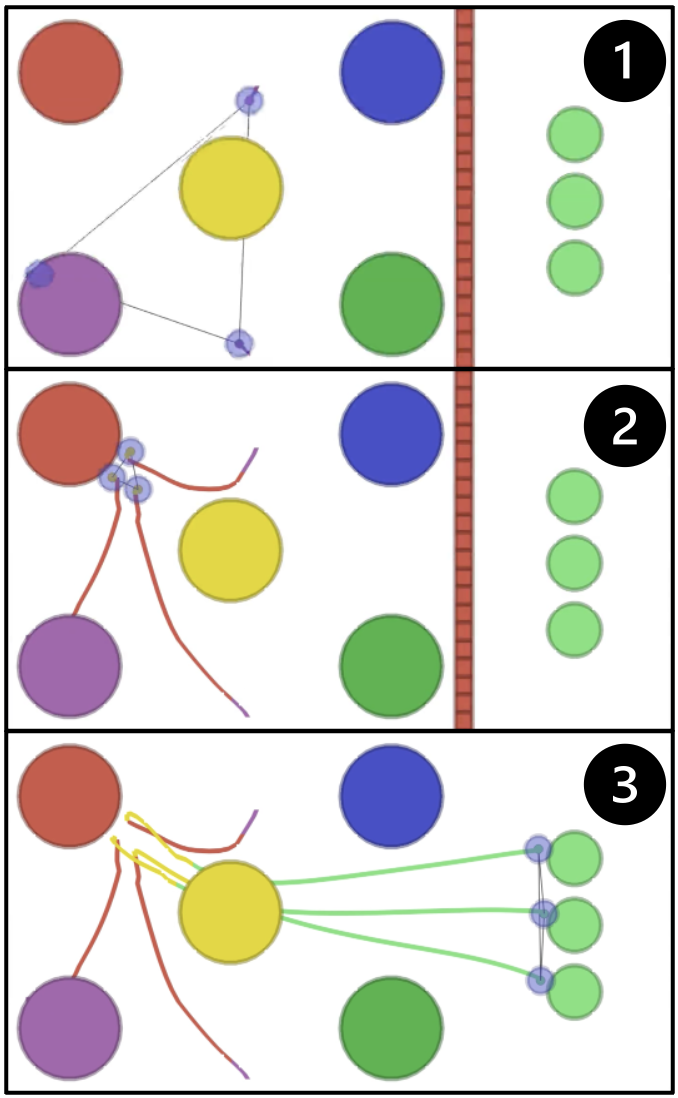}
        \caption{``Locate the purple flag, then the red flag, and proceed to the goal.''}
    \end{subfigure}\hfill
    \begin{subfigure}[t]{0.156\textwidth}
        \centering
        \includegraphics[width=\linewidth]{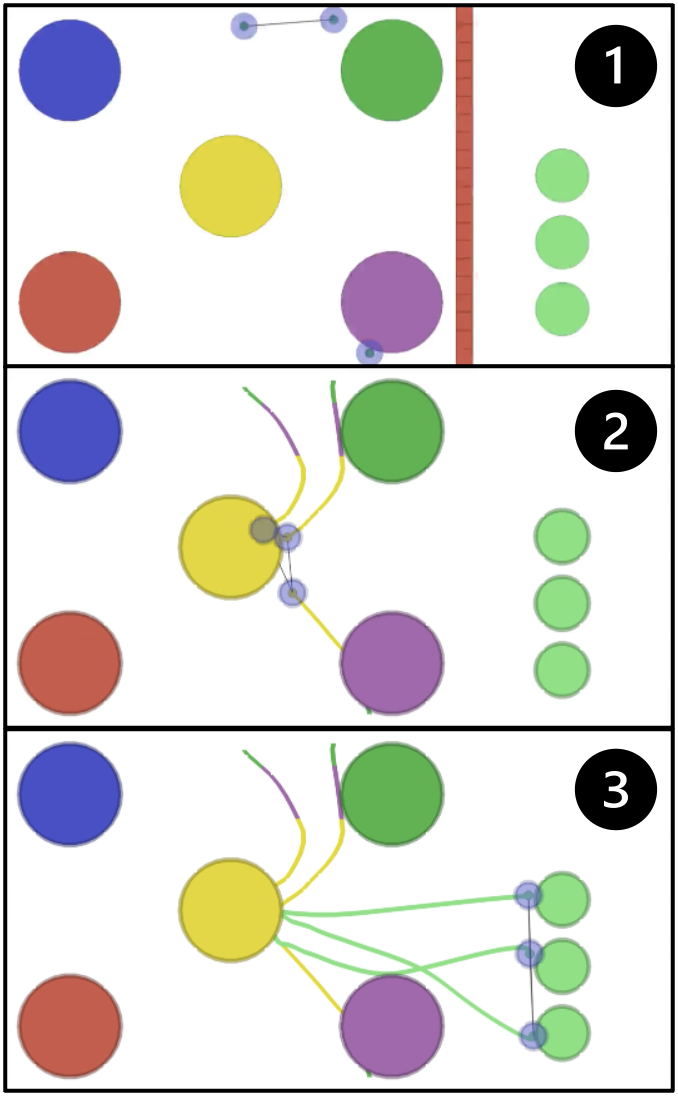}
        \caption{``Find the green flag, identify the purple one, then reach the goal.''}
    \end{subfigure}

    \caption{Examples of zero-shot multi-robot deployments in the retrieve-the-flag scenario, on physical robots (\textbf{a-c}) and in simulations (\textbf{d-f}).
    Each task requires the team to locate the specified flag, and then `bring' it to the switch (yellow) which opens a virtual gate separating their goal region.
    The team can handle a variety of natural-language expressions, objectives and even unmodeled disturbances (a person interferes with the team in (b) and (c), middle panel).
    Simulated examples add more complexity: instead of locating a single flag, the agents must find and retrieve a series of flags in a given sequence.
    \vspace{-5pt}}
    \label{fig:deployment}
\end{figure*}

Finally, we evaluate our framework for deployment on a team of robots in simulations and in real-world.
Our objective is to qualify 
robustness
against \textit{(i)} disturbances in the task space (i.e., events that may naturally occur during the course of a mission), and, \textit{(ii)} extrinsic noise in the physical world (such as what we expect in typical robot deployments).

\Cref{fig:hero} shows four snapshots from a zero-shot physical deployment using a team of three RoboMaster robots\cite{blumenkamp2024cambridge} for the task, \textit{``Identify the purple flag, navigate to the switch and proceed to the goal''}.
Each robot receives positioning data from a motion-capture system, and runs its own feedback control onboard an Nvidia Jetson Orin.
We use the ROS2 middleware to package robot policies as `nodes' and use off-the-shelf WiFi modules to exchange messages.
An offboard computer processes the input task and communicates with each robot; in general, we can distribute nodes freely across the network since our commands are not latency-critical.

In \Cref{fig:e3-realworld-evals}, we show the spatial (top-down) and temporal views from a different run of this task.
This time, we emulate a task-space disturbance, \textit{`Flag Lost'}, approx. \SI{6}{s} after the flag was found, which causes the team to double back and recover the flag before heading to the switch.
\Cref{fig:e3-realworld-evals} (top) color-codes the trajectory of one robot to indicate the team's current target (the other agents use light gray).
We observe the behavior switch as two `loops' between the purple flag and the yellow switch, corresponding to the two trips the team makes towards the flag.
The temporal view in \Cref{fig:e3-realworld-evals} (bottom) shows the process evolution. 
The robots first head towards the purple flag; upon obtaining it ($\approx\SI{3}{s}$), they change their target to the yellow switch.
At the \textit{`Flag Lost'} event ($\approx\SI{6}{s}$), the RNN resets to heading to the flag.
When the flag is found again ($\approx\SI{9}{s}$), the team continues to the switch and finally make their way to the goal.
Similar to \Cref{fig:e2-autom-rollouts}, we retain robustness against a chain of such events.

In \Cref{fig:deployment}, we show snapshots from six additional evaluations with variations on language instructions,
randomly injected \textit{`Flag Lost'} events and
unmodeled physical disruptions.
For instance, \Cref{fig:deployment}(b,c) show a person (masked for anonymity) manually displacing random robots in the team. 
{Over 8 real-robot evaluations (6 with disruptions), the team successfully recovers from different combinations of such events, thanks to both the feedback control's responsiveness and the automaton's adherence to a global task sequence.
However, under extreme physical disruptions, we observe poor tracking performance from the control policy as it also attempts to counteract the disturbance. 
This may be alleviated by adequate training and domain randomization.
}
\Cref{fig:deployment}(d--f) show evaluations {on longer task sequences}.
Similar to the multi-room problem in \Cref{subsec:eval_e0}, our method is successful due to the sub-task sequencing enforced by the automaton.

\begin{figure}[t]
    \centering
    \begin{subfigure}[t]{0.15\textwidth}
        \centering
        \includegraphics[width=\linewidth]{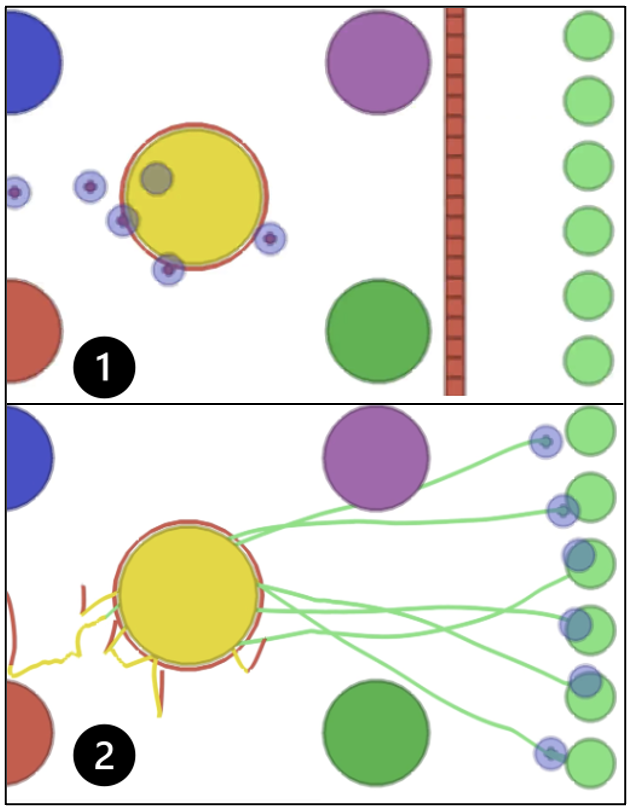}
        \caption{$\mathsf{N}\!=\!6$,\\ Inference: \SI{2.17}{ms}.}
    \end{subfigure} \hfill
    \begin{subfigure}[t]{0.15\textwidth}
        \centering
        \includegraphics[width=\linewidth]{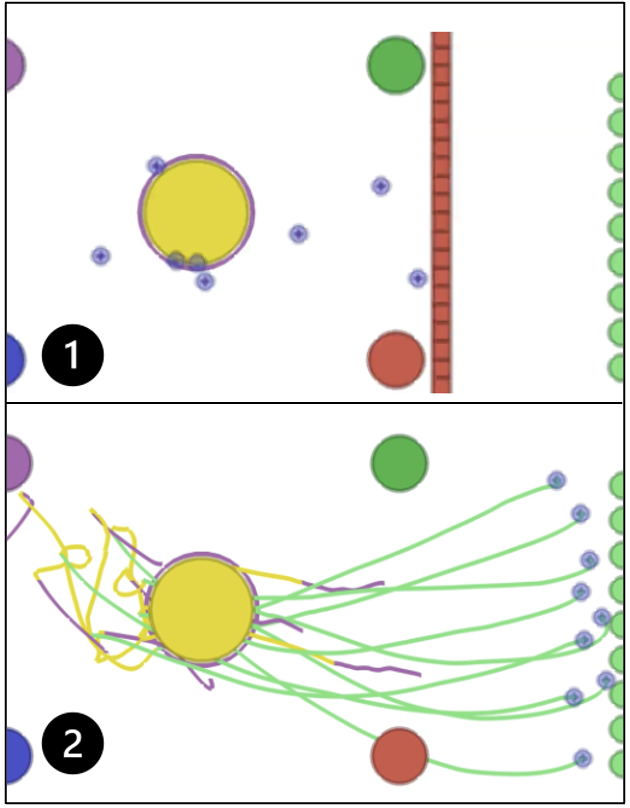}
        \caption{$\mathsf{N}\!=\!9$,\\ Inference: \SI{3.24}{ms}.}
    \end{subfigure} \hfill
    \begin{subfigure}[t]{0.15\textwidth}
        \centering
        \includegraphics[width=\linewidth]{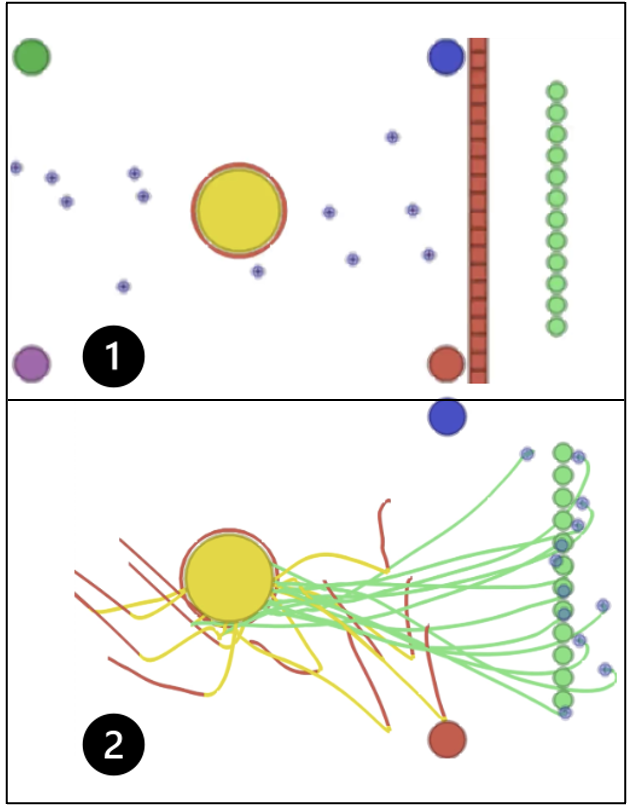}
        \caption{$\mathsf{N}\!=\!12$,\\ Inference: \SI{9.37}{ms}.}
    \end{subfigure}
    \caption{Retrieve-the-flag scenario. A policy trained on 3 robots and deployed on 6, 9, and 12 robots in simulation maintains real-time inference (measured on a MacBook Air M3).
    The task is successful,
    though the team performance drops gradually at larger disparities.
    %\vspace{-2pt}
    }
    \label{fig:n_agents}
\end{figure}

Finally, we evaluate scaling to larger teams. 
Because the policy uses a graph-based architecture, it generalizes naturally to variable team sizes. 
\Cref{fig:n_agents} shows snapshots from the same task in deployments with 6, 9, and 12 agents in simulation, using a policy that was trained with 3 agents. 
The teams complete the tasks successfully, while the inference time (per-loop) remains fairly manageable and well within realistic constraints for feedback control (under $\SI{1}{ms}$/loop/robot).

%%%%%%%%%%%%%%%           
% CONCLUSIONS %
%%%%%%%%%%%%%%%

\section{Conclusions}\label{sec:conclusion}
In this work, we presented a novel framework that enables teams of robots to perform complex, collaborative tasks from high-level natural language commands. Our approach successfully addresses the challenges of real-time reasoning, decentralized coordination, and the interpretation of ambiguous yet expressive semantic instructions in natural language. We effectively distill the reasoning capabilities of a large language model into a single, light-weight RNN. This model, paired with a graph neural network-based policy, allows robots to reason about task sequences and collaborate on sub-tasks in a fully distributed manner. Simulations and real-world experiments demonstrated that our method empowers robot teams to display interactive, onboard reasoning, leveraging semantic cues from human language to achieve their goals.
We plan to extend this work to teams with a set of predefined robot capabilities, such that they can reason over and address all tasks solvable with those capabilities.

%%%%%%%%%%%%%%           
% REFERENCES %
%%%%%%%%%%%%%%
% \balance
{\scriptsize \bibliographystyle{IEEEtran}
\bibliography{IEEEabrv,aj-refs-lang}}

\end{document}